\documentclass[10pt,twocolumn,letterpaper]{article}

\usepackage{iccv}
\usepackage{times}
\usepackage{epsfig}
\usepackage{graphicx}
\usepackage{amsmath}
\usepackage{amssymb}
\usepackage{cuted}
\usepackage{capt-of}
\usepackage{multirow}
\usepackage{mathtools}
\usepackage[accsupp]{axessibility}  

\usepackage[breaklinks=true,bookmarks=false]{hyperref}

\iccvfinalcopy

\begin{document}

\title{Relightify: Relightable 3D Faces from a Single Image via Diffusion Models}

\author{
\hspace{-2mm}Foivos~Paraperas~Papantoniou$^{1,2}$\hspace{2mm}
Alexandros~Lattas$^{1,2}$\hspace{2mm}
Stylianos~Moschoglou$^{1,2}$\hspace{2mm}
Stefanos~Zafeiriou$^{1,2}$
\\
$^{1}$Imperial College London \quad
$^{2}$Huawei Noah’s Ark Lab
\\
{\tt\small \{f.paraperas,a.lattas,s.moschoglou,s.zafeiriou\}@imperial.ac.uk}
\vspace{-7mm}
}

\maketitle

\begin{strip}
\centering
\includegraphics[width=\textwidth, trim={0 -1.5cm 0 0}, clip]{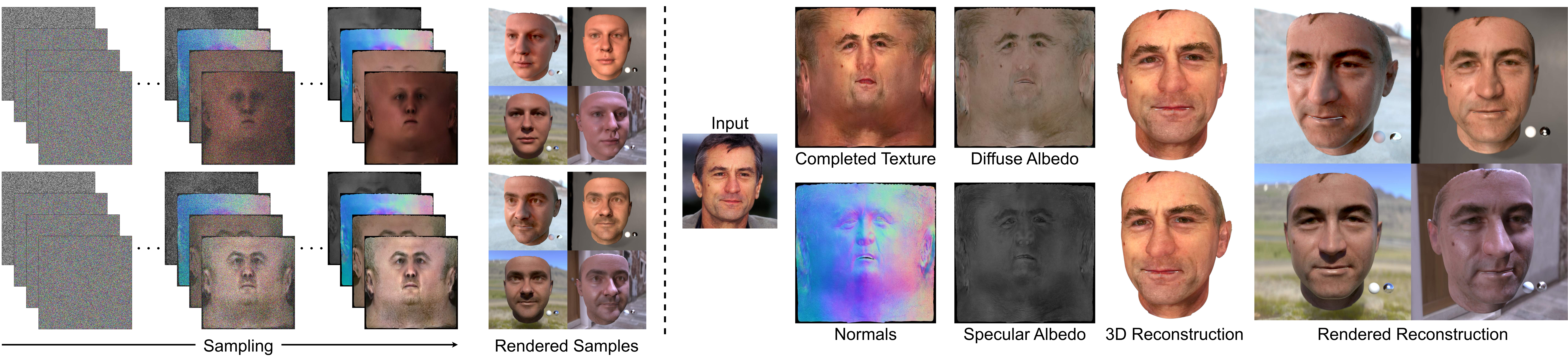}
\captionof{figure}{Left: We train an unconditional diffusion model on a high-quality dataset of UV textures and their accompanying facial reflectance maps. Right: Using this model, we perform both texture completion as well as accurate reflectance prediction from monocular images by inpainting in UV space.
Our 3D facial reconstruction requires only a single image and allows the realistic rendering of the 3D avatar.}
\label{fig:teaser}
\end{strip}

\begin{abstract}
   Following the remarkable success of diffusion models on image generation, recent works have also demonstrated their impressive ability to address a number of inverse problems in an unsupervised way, by properly constraining the sampling process based on a conditioning input.
   Motivated by this, in this paper, we present the first approach to use diffusion models as a prior for highly accurate 3D facial BRDF reconstruction from a single image. We start by leveraging a high-quality UV dataset of facial reflectance (diffuse and specular albedo and normals), which we render under varying illumination settings to simulate natural RGB textures and, then, train an unconditional diffusion model on concatenated pairs of rendered textures and reflectance components. At test time, we fit a 3D morphable model to the given image and unwrap the face in a partial UV texture. By sampling from the diffusion model, while retaining the observed texture part intact, the model inpaints not only the self-occluded areas but also the unknown reflectance components, in a single sequence of denoising steps. In contrast to existing methods, we directly acquire the observed texture from the input image, thus, resulting in more faithful and consistent reflectance estimation. Through a series of qualitative and quantitative comparisons, we demonstrate superior performance in both texture completion as well as reflectance reconstruction tasks.
   
\end{abstract}

\section{Introduction}
Creating digital avatars of real people is of paramount importance for a range of applications, including VR, AR or the film industry. Human faces have been studied extensively over the years, attracting attention at the intersection of Computer Vision, Graphics and Machine Learning research. Although vast literature exists around the estimation of the 3D shape and reflectance of a face from unconstrained inputs such as ``in-the-wild'' RGB images, it still remains a challenging problem in the field. In particular, the recent breakthrough in image synthesis using diffusion generative models creates a new perspective towards photo-realistic 3D face reconstruction, which has not been explored so far and stems from the state-of-the-art performance of these models in solving inverse problems without supervised training.

Facial reflectance capture typically requires a controllable illumination system
equipped with multiple cameras, first introduced as a Light Stage \cite{debevec2000acquiring}.
Polarized illumination and gradient patterns can be employed 
for diffuse-specular separation \cite{ma2007rapid, ghosh2011multiview},
using which, spatially varying facial reflectance maps can be acquired, that describe BRDF parameters,
including the diffuse and specular albedo and normals.
Although recent works attempt to simplify the capturing apparatus and process
using inverse rendering \cite{gotardo2018practical, riviere2020single} or commodity devices \cite{lattas2022practical},
such methods still require a laborious capturing process and expensive equipment.

Since their introduction by Blanz and Vetter~\cite{Blanz}, 3D Morphable Models (3DMMs)~\cite{BaselFaceModel, LYHM2017, FLAME:SiggraphAsia2017, FaceWarehouse, LSFM} have been established as a robust methodology for monocular 3D face reconstruction~\cite{egger20203d, 3D_survey_sota} by regularizing the otherwise ill-posed optimization problem towards a known statistical prior of the facial geometry, which is usually defined by the linear space of a PCA model. In addition to the coarse geometry estimation, 3DMMs have been used in conjunction with powerful CNN-based texture models, leading to impressively detailed avatar reconstructions even from low-resolution images~\cite{Saito_2017_CVPR, ganfit, fast-ganfit}. Furthermore, another line of research~\cite{chen2019photo, Huynh_2018_Mesoscopic, Yamaguchi2018, hifi3dface2021tencentailab, dib2021towards, dib2021practical, lattas2020avatarme, lattas2021avatarme++} revolves around the reconstruction of rendering assets such as reflectance components (diffuse and specular albedo) and high-frequency normals of the facial surface. 
As a result, the recovered 3D faces can be realistically rendered in arbitrary illumination environments.
However, prior work either contains scene illumination inhibiting relighting \cite{Deng_UVGAN,Gecer_ostec,ganfit}
or is restricted by the models' generalization, lowering the identity similarity \cite{ganfit,lattas2020avatarme,luo2021normalized}.
Our work shares the same objective in that we couple a 3DMM with high-quality UV reflectance maps,
but attempts to solve both of these issues,
by preserving the observed texture details from the input image and jointly inferring the facial reflectance.

In fact, the visible pixels of the facial texture by the given camera pose are directly recoverable from the input image via inverse rasterization of the fitted 3D mesh. Therefore, we cast the 3D face reconstruction problem as an image inpainting task in the UV space; \ie the goal is to fill in the missing pixels in a consistent manner with respect to some statistical prior. In particular, we propose to use a diffusion model as the generative backbone of our method. Diffusion models~\cite{diffusion2015} are naturally associated with guided image synthesis since they treat image generation as a sequence of denoising steps in the form of a learnable Markov process. This allows to directly interfere with the sampling process, given that samples at each part of the chain are distorted versions of real images with known noise variances. Thus, by properly modifying the sampling process, a single unconditional diffusion model can be used for different inverse problems, such as image editing~\cite{meng2022sdedit}, inpainting~\cite{RePaint_2022_CVPR, chung2022MCG}, restoration~\cite{kawar2022ddrm} or super-resolution~\cite{Come_Closer_2022_CVPR, ILVR_2021_ICCV}, without problem-specific training.  

In this paper, we build a high-quality statistical model of facial texture and reflectance by means of a diffusion model and adopt an inpainting approach to complete the partially reconstructed UV texture produced by a 3DMM fitting step. We further extend the sampling process to recover the missing reflectance components by enforcing consistency with the input texture. As a result, our method, dubbed \textit{Relightify}, generates accurate and render-ready 3D faces from unconstrained images, as shown in Fig.~\ref{fig:teaser}.

In summary, we make the following contributions:
\begin{itemize}
    \item We present the first, to the best of our knowledge, diffusion-based approach for relightable 3D face reconstruction from images. By training on a pseudo-ground-truth dataset of facial reflectance, while directly recovering texture parts from the input, we achieve high-quality rendering assets that preserve important details of the input face (\eg wrinkles, moles).
    \item We propose an efficient way of predicting different modalities in a consistent way by learning a generative model on concatenated reflectance maps and casting the reconstruction as an inpainting problem,
    spatially, but also channel-wise.
    \item We qualitatively and quantitatively demonstrate the superiority of our approach against previous methods regarding both the completed textures as well as the recovered reflectance maps.
\end{itemize}

\section{Related Work}
\subsection{Diffusion Models for Inverse Problems}
Diffusion models~\cite{diffusion2015} are latent variable generative models which artificially corrupt the data distribution by adding noise and attempt to approximate the reverse process. 
They have lately emerged as a powerful image synthesis model~\cite{ddpm_Ho, diffusion_beats_gans,song2021Score-SDE} outperforming previous state-of-the-art approaches in both conditional and unconditional tasks. While they achieve excellent image quality and are robust to multi-modal distributions, they are computationally demanding to sample from, since they require a large sequence of denoising steps (\eg 1000), each of which operates in the high dimensional image space. To alleviate this, a number of works~\cite{ddim, kong2021fast_sampling, robin2021noise_estimation} have proposed alternative strategies to accelerate sampling by reducing the steps of the reverse process. Another line of research~\cite{vahdat2021score,sinha2021d2c} proposes to train an encoding model and learn a diffusion model on its lower-dimensional latent space. Recently, Rombach \etal \cite{ldm_2022_CVPR} have further explored the use of a VQGAN~\cite{Esser_2021_VQGAN} as the auto-encoding model, showing that a mild compression is enough to reduce the training/sampling time without sacrificing sample quality. The latter approach is our method of choice for this work, as we elaborate on a high-resolution UV image space, which would otherwise significantly increase the computational overhead.

One of the most interesting aspects of diffusion models is that they can be used as unsupervised solvers for different inverse problems, where the goal is to reconstruct a sample from some distorted observation, \ie conditioning input. Song \etal \cite{song2021Score-SDE} propose a conditioning mechanism during inference that allows applications such as class-conditional generation, inpainting and colorization. Similarly, \cite{ILVR_2021_ICCV} uses a low-pass filtered version of the conditioning image to guide the denoising process at each step and SDEdit~\cite{meng2022sdedit} addresses image translation and editing using a diffused version of the input image to initialize sampling from an intermediate timestep. RePaint~\cite{RePaint_2022_CVPR} achieves state-of-the-art results on image inpainting by repeating multiple forward and backward diffusion steps to enforce harmonization. Despite its improved performance, this resampling strategy significantly increases the computational time. In contrast, CCDF~\cite{Come_Closer_2022_CVPR} and DDRM~\cite{kawar2022ddrm} propose efficient techniques for reducing the length of the reverse process while retaining image quality at a high level. More recently, MCG~\cite{chung2022MCG} introduced a novel manifold constraint step, which combined with the standard reverse diffusion outperforms the aforementioned methods on a number of inverse tasks, including inpainting. We adopt this approach in our work to accurately fill in the missing pixels of both texture and reflectance maps of a face from a given image via diffusion-based inpainting, while fully preserving the observed ones. Note also that this approach does not assume any specific distribution of visibility masks, as it is trained unconditionally on complete textures.

\subsection{Facial Reconstruction}
3DMMs \cite{Blanz} are the typical models for facial reconstruction from ``in-the-wild'' images,
using a linear model for the identity, and additional linear models for expression or color. Current facial 3DMMs include the Basel Face Model (BFM) \cite{BaselFaceModel}
and the Large Scale Facial Model (LSFM) \cite{LSFM}. Egger \etal \cite{egger20203d} provide a thorough review on the subject. AlbedoMM \cite{smith2020morphable} first created a 3DMM of facial reflectance, which can be relighted,
but is restricted to a linear and per-vertex color model. Dib \etal~\cite{dib2021practical,dib2021towards} improved on prior works' simplistic shading models and used inverse ray tracing to acquire photorealistic facial reflectance. Recently, GANFit \cite{ganfit,fast-ganfit} introduced a potent method for fitting 3DMMs with a GAN-based \cite{goodfellow2020generative} facial texture generator, achieving high-fidelity facial avatars, but lacking relighting capabilities due to baked illumination in the textures. AvatarMe++ \cite{lattas2020avatarme, lattas2021avatarme++} overcame this issue by
translating the reconstructed textures to facial reflectance using a conditional GAN, while adding extra processing steps.
While we use AvatarMe++ to augment our training data,
our method significantly outperforms them by using a powerful diffusion model and inferring only the occluded facial texture.

TBGAN \cite{gecer2020synthesizing} first introduced a deep generative network 
for facial reflectance, based on ProgressiveGAN \cite{karrasprogressive}
and \cite{li2020learning} introduced a more powerful model, based on StyleGAN \cite{karras2019style}. However, both works did not showcase fitting capabilities. An extension of the latter \cite{luo2021normalized},
introduced a set of multiple networks, with a StyleGAN2 \cite{karras2020analyzing} base,
that can be used to generate shape and albedo from images with arbitrary illumination and expression.
While close to our work,
our method uses a single and more powerful diffusion model,
inferring not only the diffuse albedo, but also the specular albedo and normals.
Moreover, our work inpaints only the occluded facial areas, preserving the visible part of the texture and achieves higher reconstruction fidelity. 

Although our method is applied to facial reconstruction,
we simultaneously solve a facial texture inpainting problem in UV space. Initially explored in 2D facial images \cite{li2017generative}
and expanded to UV completion using deep encoder-decoder architectures (UV-GAN \cite{Deng_UVGAN}),
such works recover the facial texture from partial and masked facial images.
Recently, OSTeC \cite{Gecer_ostec}, used a pre-trained StyleGAN
in 2D to recover multiple poses of the input subject so as to create a complete UV facial texture.
While prior works achieve impressive results,
all are restricted facial textures with baked illumination.
In contrast, we jointly recover the facial reflectance,
making the reconstruction relightable in standard rendering engines.

\section{Method}
\begin{figure*}[h]
\centering
\includegraphics[width=1.0\textwidth]{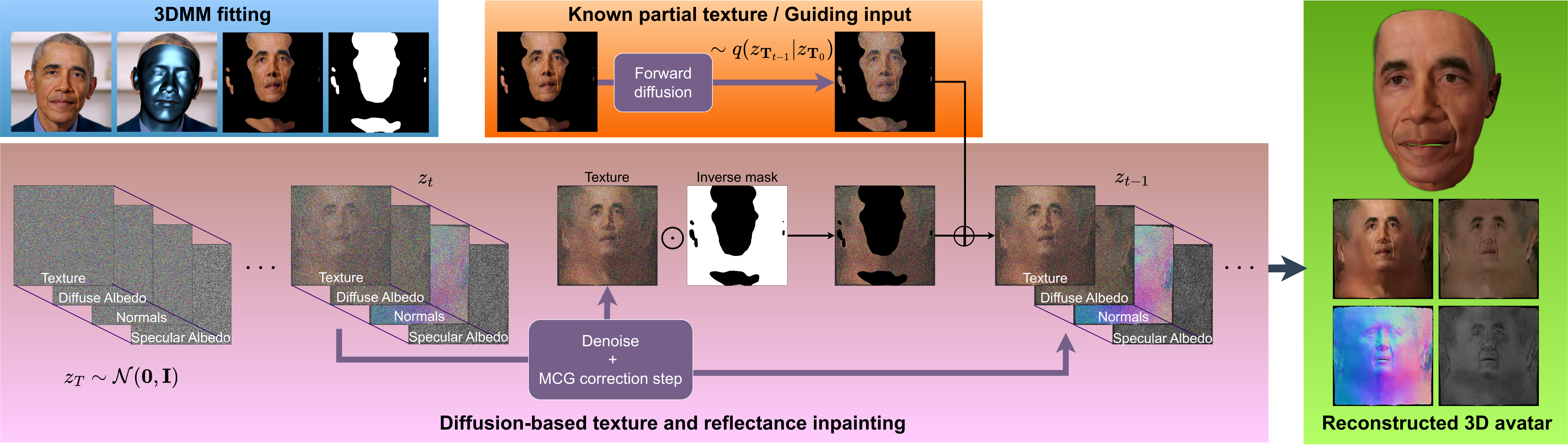}
\caption{Overview of our method during inference. Please note that we use a latent diffusion model~\cite{ldm_2022_CVPR}, yet we illustrate the denoising process in the original image space for visualization purposes. We perform standard 3DMM fitting to get a partial UV texture via image-to-uv rasterization. Then, starting from random noise, we utilize the known texture to guide the sampling process of a texture/reflectance diffusion model towards completing the unobserved pixels. Each denoising step, from $z_t$ to $z_{t-1}$ ($t\in\{1,\ldots,T\}$), follows an inpainting approach similar to MCG~\cite{chung2022MCG} (see Eq.~\ref{eq:inpainting_steps}): 1) The reflectance maps and unobserved texture pixels are updated based on reverse diffusion sampling and manifold constraints, while 2) the known pixels are directly sampled from the input texture via forward diffusion ($\odot$ and $\oplus$ denote the Hadamard product and addition respectively). Note that masking is only applied to the texture, while the reflectance maps (diffuse/specular albedo, normals) are entirely predicted from random noise. At the end of the process, we acquire high-quality rendering assets, making our 3D avatar realistically renderable.}
\label{fig:method}
\end{figure*}

\begin{figure}[h]
\centering
\includegraphics[width=\linewidth]{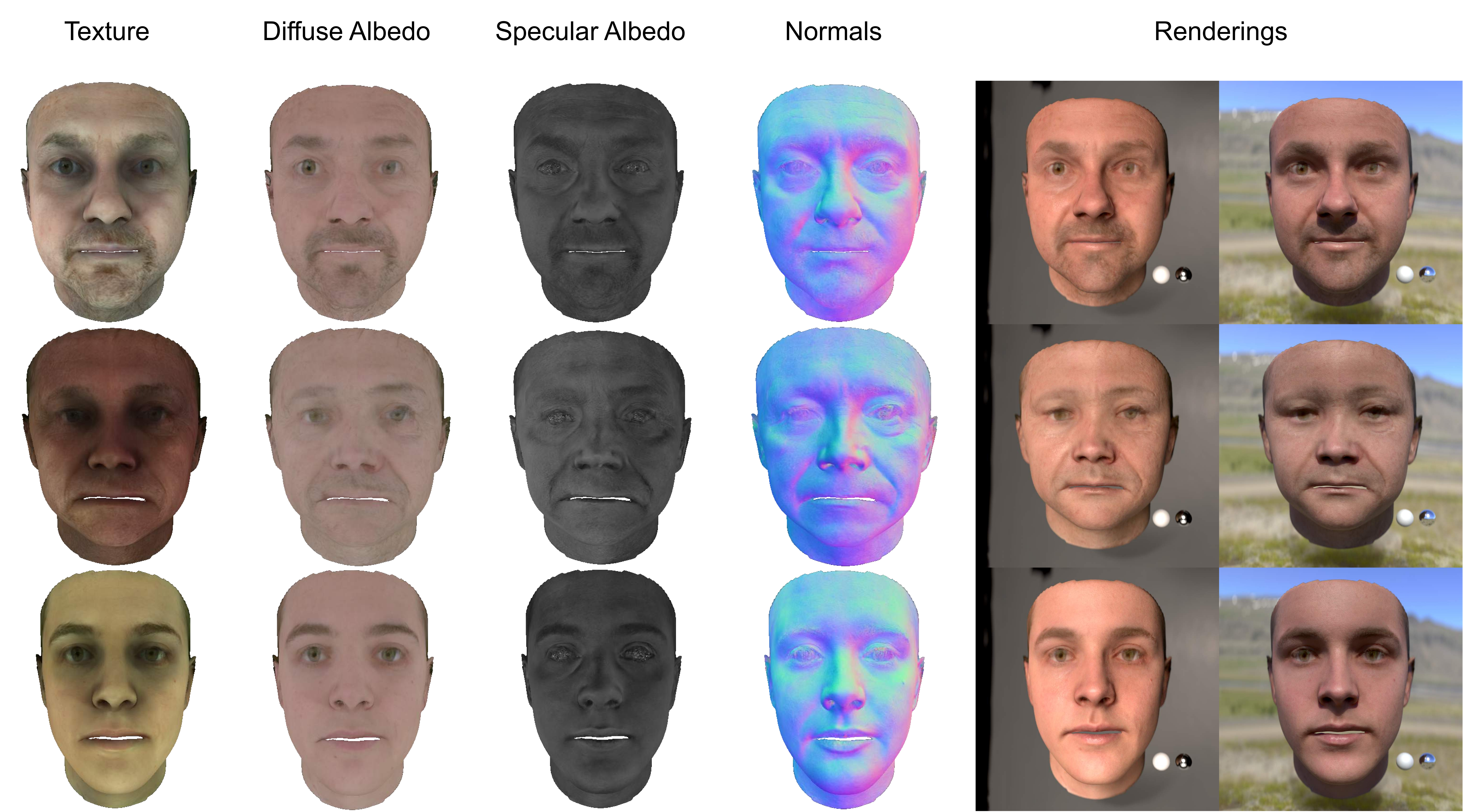}
\caption{Samples of texture and reflectance maps generated by our diffusion model (left) and renderings in different scenes (right). Random 3DMM shapes are used for the visualization.}
\label{fig:samples}
\end{figure}

\begin{figure*}[h!]
\centering
\includegraphics[width=1.0\textwidth]{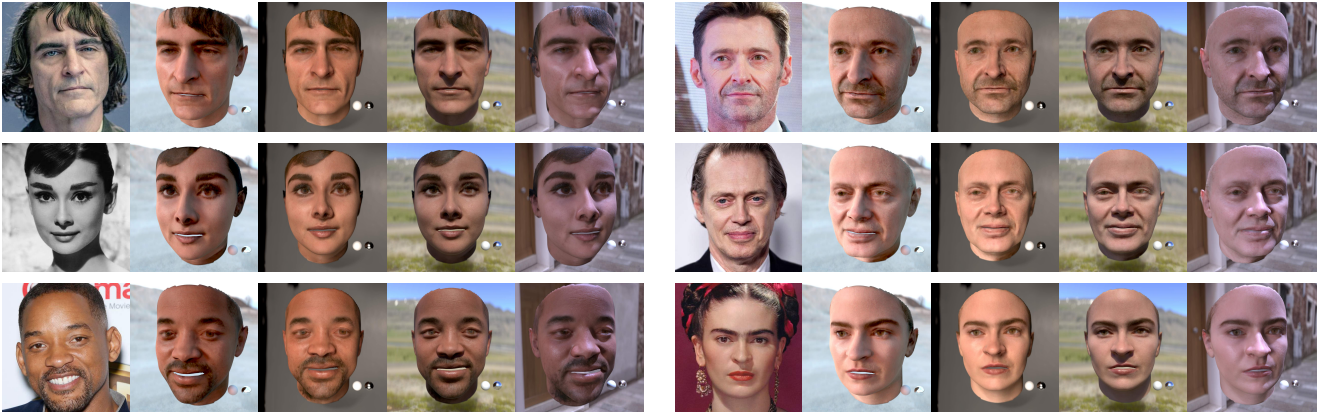}
\caption{Examples of 3D reconstructions by our method, rendered using different environment maps in a commercial renderer \cite{marmoset}.}
\label{fig:renders}
\end{figure*}

We propose a diffusion-based inpainting approach to estimate both the UV texture with existing baked illumination and the actual reflectance of a face in a single process. At the core of our approach lies an unconditional diffusion generative model trained on pairs of textures and their accompanying reflectance. This coupled texture-reflectance modeling along with the sequential denoising process of diffusion models allows us to reconstruct the reflectance from a partial texture of the input face, as shown in Fig.~\ref{fig:method}. Our method, thus, generates high-quality 3D face avatars from ``in-the-wild'' images, which can be realistically relighted.

In the following sections, we first analyze the training of our diffusion model, and then explain the 3D shape reconstruction and texture inpainting strategies in further detail.

\subsection{Diffusion Models: Background}
\label{sub:diffusion_models}
Given a distribution of real images $x$, diffusion models~\cite{diffusion2015} define a forward diffusion process which gradually adds Gaussian noise to the input image in $T$ consecutive steps. This corresponds to a fixed Markov Chain, where starting from a clean image $x_0$, the noisy samples $x_t$ at each timestep $t$ are drawn from the following distributions (with timestep-depending variances $\beta_t$) conditioned on the previous samples:
\begin{equation}
    q(x_t | x_{t-1}) = \mathcal{N}(x_t; \sqrt{1-\beta_t} x_{t-1}, \beta_t \mathbf{I})
\end{equation}
This is equivalent to directly sampling $x_t$ conditioned on the clean image $x_0$ via:
\begin{equation}
    q(x_t | x_0) = \mathcal{N}(x_t; \sqrt{\bar{\alpha}_t} x_0, (1-\bar{\alpha}_t) \mathbf{I})
\end{equation}
where $\alpha_t \coloneqq 1-\beta_t$ and $\bar\alpha_t \coloneqq \prod_{s=1}^t \alpha_s$.
Given large enough $T$, this process leads to normally distributed noise $x_T$. Then, the goal is to learn the reverse Markov process:
\begin{equation}
p_{\theta}(x_{t-1}|x_t) = \mathcal{N}(x_{t-1}; \mu_{\theta}(x_t, t), \Sigma_{\theta}(x_t, t))
\end{equation}
which gradually denoises the random noise $x_T$ towards a realistic image, by minimizing the variational bound of the negative log likelihood~\cite{ddpm_Ho,diffusion_beats_gans}. Following the reparameterization proposed in~\cite{ddpm_Ho}, the model consists of time-conditioned denoising autoencoders $\epsilon_{\theta}(x_t, t); t\in\{1,2,\ldots,T\}$, which are trained to predict the noise $\epsilon\sim\mathcal{N}(\mathbf{0},\mathbf{I})$ that was added to the input image $x_0$ to account for the noisy version $x_t$:
\begin{equation}
    L = E_{x_0,\epsilon,t}\left[ || \epsilon - \epsilon_{\theta}(x_t, t) ||^2 \right]
\end{equation}
Once trained, we can generate images by starting from random noise $x_T\sim\mathcal{N}(\mathbf{0},\mathbf{I})$ and sequentially drawing denoised images around the mean:
\begin{equation}
\mu_{\theta}(x_t, t) = \frac{1}{\sqrt{\alpha_t}} \left( x_t - \frac{\beta_t}{\sqrt{1-\bar{\alpha}_t}} \epsilon_{\theta}(x_t, t) \right)
\end{equation}

\subsection{Training of our Diffusion Model}
In this work, we harness the power of diffusion models to learn a strong generative prior over the domain of facial texture/reflectance. In particular, we adopt a physically-based perspective by separating the facial reflectance into different UV maps, namely diffuse albedo ($\mathbf{A}_d$), specular albedo ($\mathbf{A}_s$) and surface normals ($\mathbf{N}$) with high-frequency details. This allows realistic rendering under different illumination conditions. We learn our prior using a high-quality dataset consisting of \emph{complete} pairs of facial reflectance, and a corresponding rendered texture $\mathbf{T}$ under arbitrary illumination. More details on the data we use are provided in section~\ref{dataset}. We train an unconditional diffusion model (as described in section \ref{sub:diffusion_models}) on the quadruples:
\begin{equation}
   x = \left[\mathbf{T},\mathbf{A}_d,\mathbf{A}_s,\mathbf{N}\right]\in\mathbb{R}^{512 \times 512 \times 10}\label{eq:quad}
\end{equation}
where we concatenate the components of Eq.~\ref{eq:quad} across channels (each of the 4 UV images measures ${512 \times 512}$ pixels and 3 channels, except for the single-channel image $\mathbf{A}_d$). By sampling from this model, we can synthesize pairs of shaded RGB textures ($\mathbf{T}$) and reflectance components ($\mathbf{A}_d, \mathbf{A}_s, \mathbf{N}$) which are in correspondence, meaning that the texture is a rendered version of the UV reflectance under some illumination environment.

In practice, to reduce the computational requirements to a reasonable level, we follow the paradigm of \textbf{latent diffusion models} proposed by Rombach \etal~\cite{ldm_2022_CVPR}, where the images are first compressed to a latent space $z=\mathcal{E}(x)\in\mathbb{R}^{h \times w \times c}$ by training a perceptual auto-encoder, consisting of an encoder $\mathcal{E}$ and a decoder $\mathcal{D}$. Using perceptual and adversarial losses similar to VQGAN~\cite{Esser_2021_VQGAN}, the autoencoder achieves an excellent quality of reconstructed samples $\tilde{x}=\mathcal{D}(\mathcal{E}(x))$, while allowing to efficiently train the diffusion model on the lower dimensional pixel-space of the learned embeddings. In our case, we train four similar auto-encoders, one for each of $\mathbf{T}, \mathbf{A}_d, \mathbf{A}_s$ and $\mathbf{N}$, all of them reducing the input resolution to latent dimensions of $h=w=64$, $c=3$. Therefore, our latent diffusion model~\cite{ldm_2022_CVPR} is trained on the concatenation of the 4 embeddings:
\begin{equation}
    z = \left[z_{\mathbf{T}},z_{\mathbf{A}_d}, z_{\mathbf{A}_s}, z_{\mathbf{N}}\right]\in\mathbb{R}^{64 \times 64 \times 12}
\end{equation}
Samples from our diffusion model (after being decoded through each $\mathcal{D}$) can be seen in Fig.~\ref{fig:teaser} (left part) and Fig.~\ref{fig:samples}.

\subsection{Inference}
We use the aforementioned trained diffusion model to perform inpainting on both the texture and reflectance UV maps based on a partial UV texture obtained by 3DMM fitting. We provide a detailed description below.

\paragraph{3DMM Fitting and Texture Initialization.}
We rely on 3DMMs to recover a rough 3D shape of the face from a 2D image as a mesh $\mathbf{S}\in\mathbb{R}^{n \times 3}$ with $n$ vertices. Specifically, we employ a linear 3DMM:
\begin{equation}\
    \mathbf{S}(\mathbf{p}_s, \mathbf{p}_e)=\mathbf{m}+\mathbf{U}_s\mathbf{p}_s+\mathbf{U}_e\mathbf{p}_e
\end{equation}
consisting of the LSFM~\cite{LSFM} shape eigenbasis $\mathbf{U}_s\in\mathbb{R}^{3n \times 158}$ and the expression eigenbasis $\mathbf{U}_e\in\mathbb{R}^{3n \times 29}$ from the 4DFAB database~\cite{4DFAB}. We fit the 3DMM to the input image by optimizing the shape coefficients $\mathbf{p}_s$, expression coefficients $\mathbf{p}_e$ and camera parameters $\mathbf{p}_c$ using an off-the-shelf framework \footnote{\url{https://github.com/ascust/3DMM-Fitting-Pytorch}}. Note that any 3DMM fitting framework works as a ``plug and play'' solution to our method. Thus, one may trivially use a more sophisticated algorithm (\eg GANFit~\cite{ganfit}) for precise shape reconstruction.

We use a standard UV topology for texturing the 3D mesh, where each vertex is assigned to a fixed 2D coordinate on the UV plane. By rasterizing the fitted 3D mesh and using barycentric interpolation, we can reverse the rendering process and unfold the face in UV, hence reconstructing the visible parts of the texture directly from the input image. This initial texture is accompanied by a UV visibility mask, with 1 for pixels that are observed from the input image, and 0 for those that are occluded and, thus, need to be inpainted by our model.

\paragraph{Texture Completion and Reflectance Prediction.}
Starting from the partially completed UV texture $\mathbf{T}_0$ of the face and a binary visibility mask $m$ produced by the previous step, our goal is to inpaint the remaining pixels along with the pixels of the 3 reflectance maps. We use the latent representation $z_{\mathbf{T}_0} = \mathcal{E}(\mathbf{T}_0)\in\mathbb{R}^{h \times w \times c}$ of this texture image to constrain the reverse diffusion process. Note that the mask $m$ is downsampled to the same resolution $h=w=64$ of the latent space for the next steps. Our inpainting algorithm starts with a random noise image $z_T\sim\mathcal{N}(\mathbf{0},\mathbf{I})$ and uses the denoising procedure of MCG~\cite{chung2022MCG}, consisting of the following repeated steps:
\begin{subequations}
\begin{align}
    z_{t-1}^\text{unknown} &\sim \mathcal{N}(\mu_{\theta}(z_t, t), \Sigma_{\theta}(z_t, t)) \label{eq:reverse_diff}\\
    z_{\mathbf{T}_{t-1}}^\text{known} &\sim \mathcal{N}(\sqrt{\bar{\alpha}_{t-1}} z_{\mathbf{T}_0}, (1-\bar{\alpha}_{t-1}) \mathbf{I}) \label{eq:forward_diff}\\
    \hat{z}_0 &= \left( z_t - \sqrt{1-\bar\alpha_t}\epsilon_\theta(z_t, t) \right) / \sqrt{\bar\alpha_t} \label{eq:pred_x0}\\
    \mathcal{L} &= \|\left( z_{\mathbf{T}_0} - \hat{z}_{\mathbf{T}_0} \right)\odot m\|_2^2 \label{eq:mcgloss}\\
   z_{\mathbf{T}_{t-1}} &= m \odot z_{\mathbf{T}_{t-1}}^\text{known} + (1-m) \odot \left( z_{\mathbf{T}_{t-1}}^\text{unknown} -\alpha\frac{\partial \mathcal{L}}{\partial z_{\mathbf{T}_t}} \right) \label{eq:update_tex}\\
   z_{k_{t-1}} &= z_{k_{t-1}}^\text{unknown} -\alpha\frac{\partial \mathcal{L}}{\partial z_{k_t}},\quad k=\{\mathbf{A}_d, \mathbf{A}_s, \mathbf{N}\} \label{eq:update_brdf}
\end{align}
\label{eq:inpainting_steps}
\end{subequations}
Given a sample $z_t$ at timestep $t$, we first sample the next denoised sample $z_{t-1}$ using the original reverse diffusion step (Eq.~\ref{eq:reverse_diff}). We term this as $z_{t-1}^\text{unknown}$ (borrowing the notation from~\cite{RePaint_2022_CVPR}) as it does not take into account the known parts of the observed texture. To exploit the known texture, we sample a noisy version of it $z_{\mathbf{T}_{t-1}}^\text{known}$ at timestep $t-1$ via a forward diffusion step (Eq.~\ref{eq:forward_diff}). Then, we directly impose this known noisy texture $m \odot z_{\mathbf{T}_{t-1}}^\text{known}$ ($\odot$ denotes the Hadamard product) as in the first half of Eq.~\ref{eq:update_tex}. Finally, for the unknown pixels, we add the manifold constraint introduced in MCG~\cite{chung2022MCG}; \ie we make a prediction of the clean sample $\hat{z}_0$ (Eq.~\ref{eq:pred_x0}) based on the previous timestep $z_t$, compare this ($\ell_2$ loss) with the ground truth in the known regions (Eq.~\ref{eq:mcgloss}), and use the gradient of this loss to update the unknown pixels of $z_{t-1}$ (Eq.~\ref{eq:update_tex} and \ref{eq:update_brdf}) so as to minimize this distance.

\paragraph{Note on inpainting algorithm.} We have chosen to adopt the recently proposed MCG~\cite{chung2022MCG} inpainting algorithm, which outperforms related state-of-the-art diffusion-based methods (\eg RePaint~\cite{RePaint_2022_CVPR}, DDRM~\cite{kawar2022ddrm}), as we empirically found it to produce excellent results. Motivated by the original algorithm, which aims at inpainting standard RGB images, we expand it to account for different input domains: by treating our images as concatenated texture/reflectance maps, we force the model to perform not only spatial inpainting, but also ``channel-wise inpainting'', by filling the missing pixels in a manner that closely aligns with the training distribution. This essentially encourages the model to learn an inverse rendering transformation during testing, thus predicting accurate reflectance maps from just a partial illuminated version of them, despite not directly imposing physically-based constraints.

\begin{figure}[h]
\centering
\includegraphics[width=.98\linewidth]{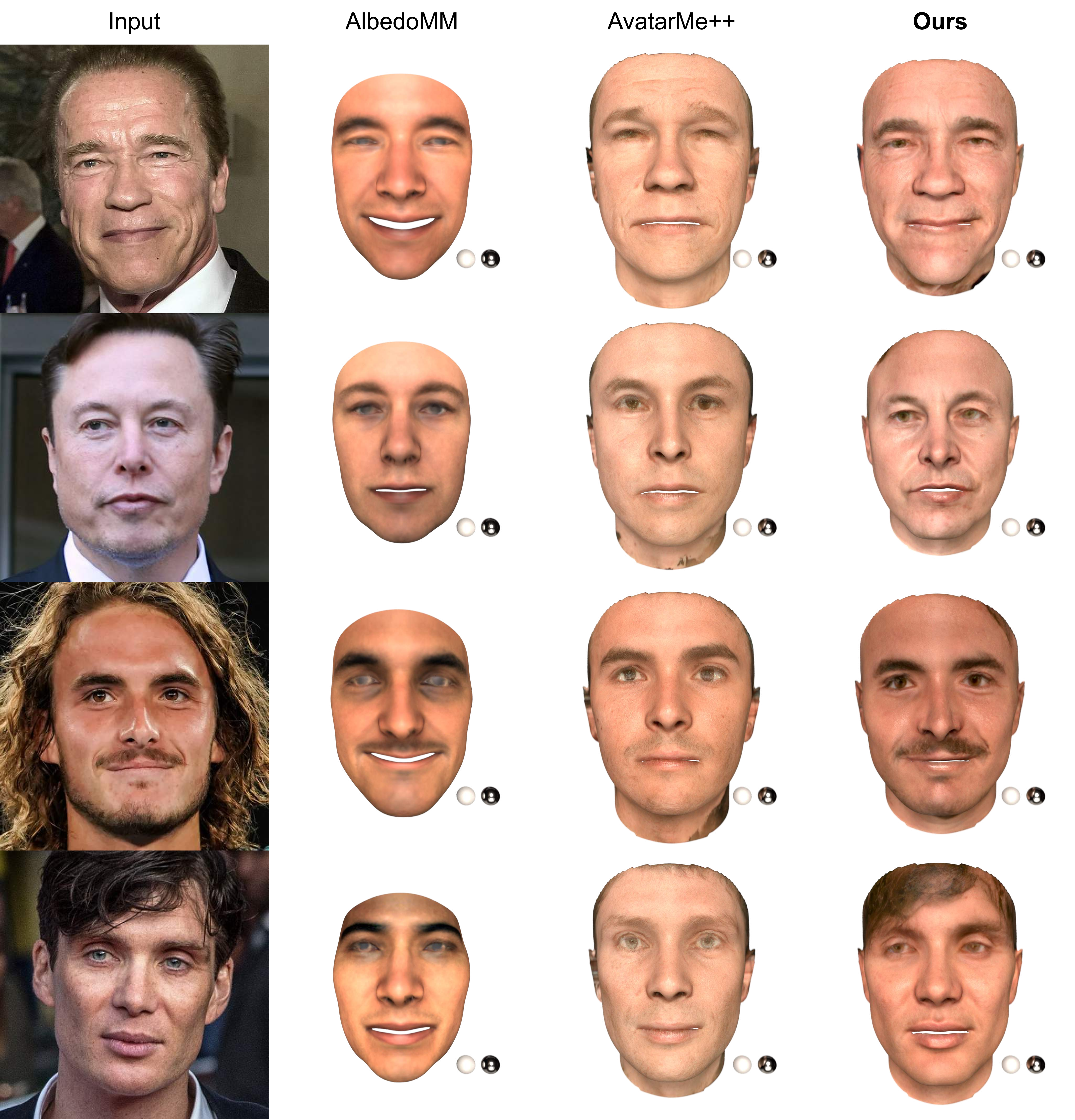}\\
\caption{
Rendered reconstructions of shape and reflectance
by AlbedoMM \cite{smith2020morphable} (using the open-source code), AvatarMe++ \cite{lattas2021avatarme++} (provided by authors) and our method,
in the same illumination.}
\label{fig:relightify_vs_amm_vs_av}
\end{figure}

\section{Experiments}

\begin{figure}[h]
\centering
\includegraphics[width=.48\textwidth]{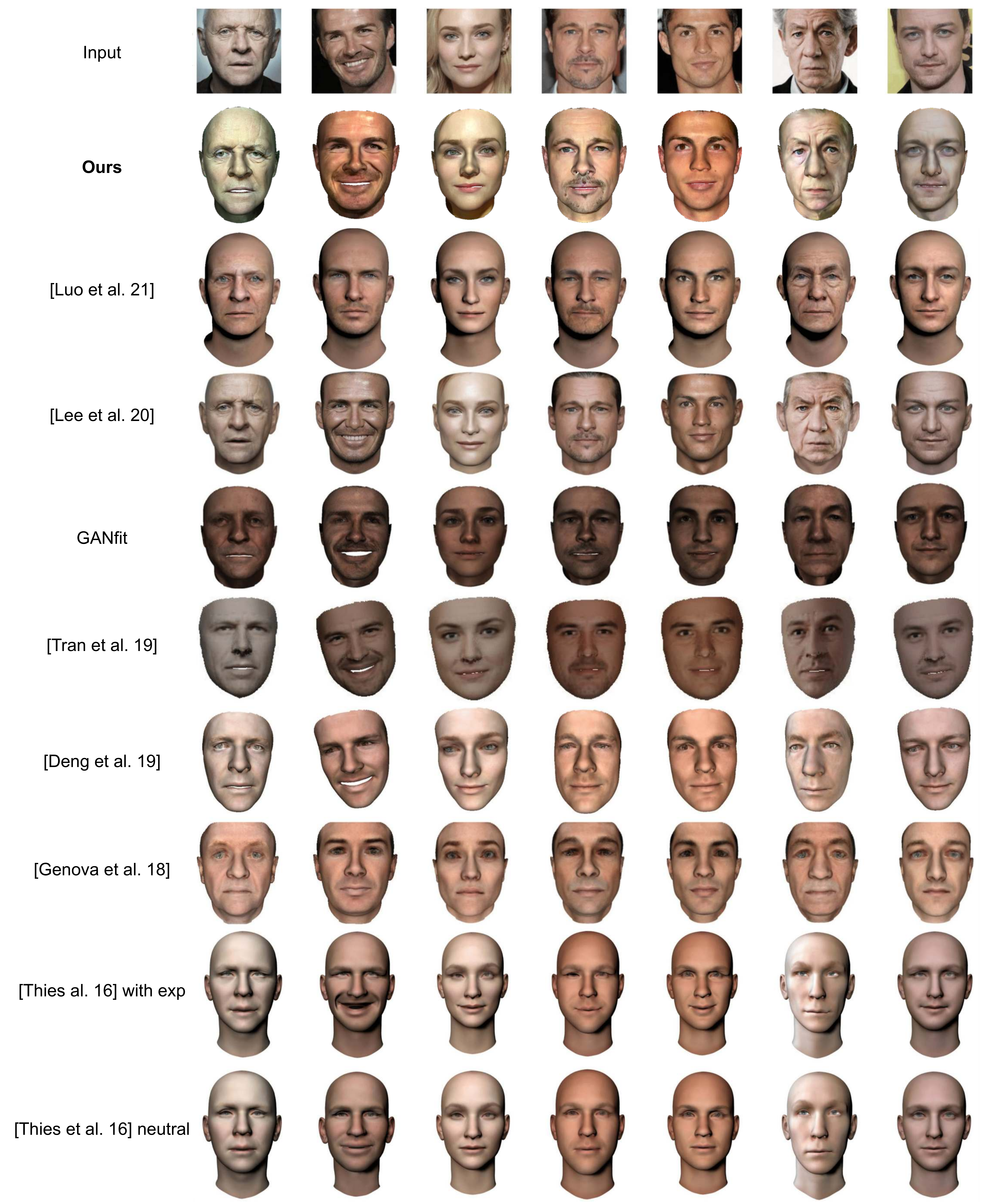}\\
\caption{Visual comparison with state-of-the-art 3D face reconstruction methods~\cite{luo2021normalized, Lee_2020_CVPR, ganfit, Tran_2019_CVPR, deng2019accurate, genova2018unsupervised, Thies_2016_face2face}. Results for related methods are borrowed from~\cite{luo2021normalized}.}
\label{fig:3D_recon_comparison}
\end{figure}

\begin{figure}[h]
    \centering
    \includegraphics[width=\linewidth]{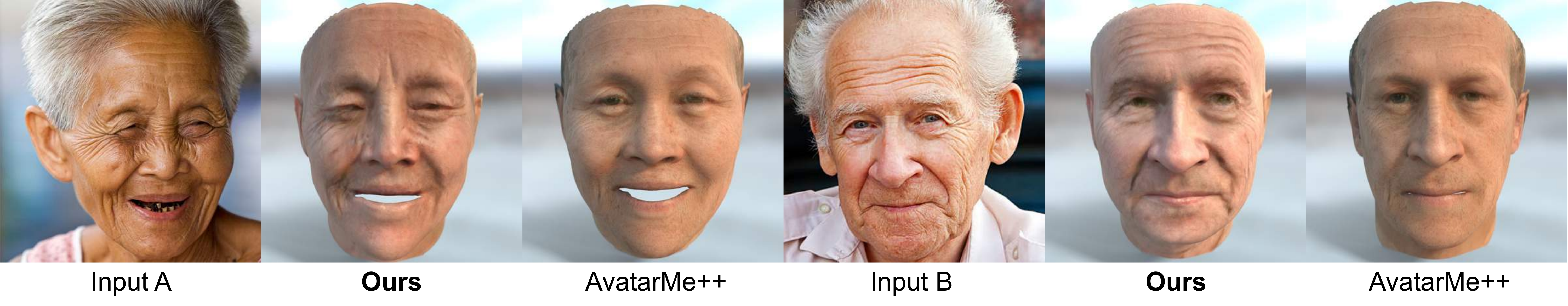}
    \caption{Comparison with AvatarMe++ \cite{lattas2021avatarme++} challenging cases.}
    \label{fig:relightify_vs_av}
    \vspace{-1mm}
\end{figure}

\subsection{Dataset and Implementation Details}
\label{dataset}
We create a high-quality dataset that consists of facial textures 
and their corresponding reflectance. Each item includes 
a texture $\mathbf{T}$, shaded in some illumination,
diffuse albedo $\mathbf{A}_d$, specular albedo $\mathbf{A}_s$ and normals $\mathbf{N}$.
To achieve this, firstly, we acquire the public MimicMe dataset \cite{papaioannou2022mimicme},
which contains $\mathbf{\tilde{T}} = \{\mathbf{T}_0, \dots, \mathbf{T}_{n_T}\}, n_T = 4,700$ diverse facial textures,
whose statistics are reported in \cite{papaioannou2022mimicme}.
However, such textures contain the illumination of the scanning apparatus and are not relightable.
Hence, we then train an image-to-image translation network based on AvatarMe++ model using the available dataset \cite{lattas2021avatarme++},
which translates the textures $\mathbf{\tilde{T}}$ to facial reflectance:
$\alpha(\mathbf{\tilde{T}}) \rightarrow \{\mathbf{A}_D, \mathbf{A}_S, \mathbf{N}\}$.
Moreover, we augment the skin-tone diversity,
using histogram matching albedo augmentation following \cite{lattas2023fitme}.
Given the memory requirement of our network,
all textures have a resolution of $512\times{}512$.
Finally, to enable the diffusion model to perform well in ``in-the-wild'' images, we use the shapes $\mathbf{S}$ of MimicMe and the acquired reflectance,
to re-render the textures under arbitrary realistic environments,
directly on the UV space: 
$\rho(\mathbf{A}_D, \mathbf{A}_S, \mathbf{N}, \mathbf{S}) \rightarrow \mathbf{T}$. For an evaluation of the model without re-rendered textures, please refer to the Supp.~Material.
Although AvatarMe++ uses a similar method to augment training data,
we do not require this process to be differentiable and use a ray-tracing renderer \cite{marmoset} (\textit{Baker} algorithm) to achieve more realistic textures.

To train our model, we use a KL-regularized latent diffusion model with the default hyper-parameters proposed by the authors of ~\cite{ldm_2022_CVPR}. Specifically, we use a downsampling factor of $f=8$ for the perceptual auto-encoder and a diffusion length of $T=1000$ for the denoising model. We train our model once and use it for texture and reflectance reconstruction from ``in-the-wild'' images. Below we provide comprehensive qualitative and quantitative evaluations.

\subsection{Qualitative Results}
\begin{figure}[h]
\centering
\includegraphics[width=.47\textwidth]{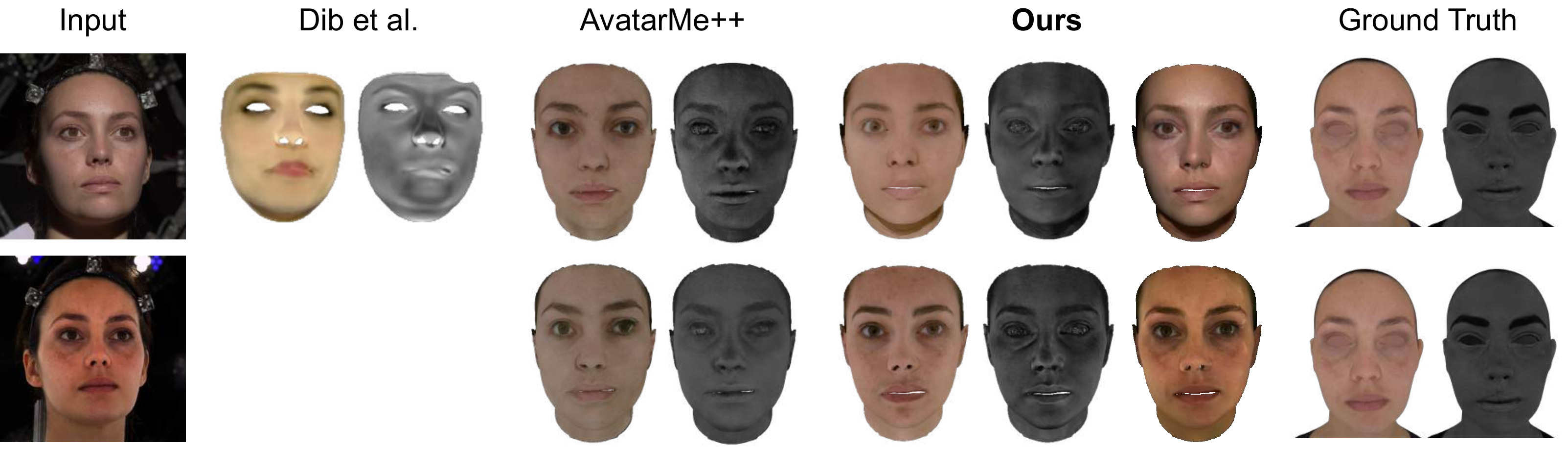}\\
\caption{Reconstructions from images with different illumination (Digital Emily Project~\cite{DigitalEmily}) by our method as well as~\cite{dib2021towards, lattas2021avatarme++} and ground truth. We show the diffuse and specular albedo for all methods (where available), plus the recovered texture for our method.}
\label{fig:emily}
\end{figure}

As already described, we produce relightable 3D faces with reflectance assets that are compatible with commercial rendering engines. Fig.~\ref{fig:renders} shows examples of reconstructions from ``in-the-wild'' images and realistic renderings in varying environments (more results are included in the Supp.~Material). Furthermore, we provide a visual comparison with the reflectance reconstruction methods of AlbedoMM~\cite{smith2020morphable} and AvatarMe++~\cite{lattas2021avatarme++} in Fig.~\ref{fig:relightify_vs_amm_vs_av}. As can be seen, we recover 3D faces of higher consistency with respect to the input. 
Note that AvatarMe++~\cite{lattas2021avatarme++} starts from a GAN-generated texture as input,
without direct feedback from the actual facial image. Despite using it to create our training data, our method clearly outperforms AvatarMe++~\cite{lattas2021avatarme++} during testing by conditioning the reflectance prediction on the genuine visible facial texture instead of a statistical approximation (fitting) of it (see Fig.~\ref{fig:relightify_vs_av} for some challenging subjects). We also show an extensive qualitative comparison with related 3D reconstruction methods in Fig.~\ref{fig:3D_recon_comparison} (most of which can only recover the texture), where similar observations can be made. Finally, we test our method on images from the Digital Emily ~\cite{DigitalEmily} and show the results in Fig.~\ref{fig:emily} together with related works~\cite{dib2021towards, lattas2021avatarme++}. We yield similar results regardless of the lighting, thanks to our coupled texture/reflectance modeling that combines reflectance with randomly rendered textures during training.

\subsection{Texture Completion}

\begin{figure}[h]
\centering
\includegraphics[width=.47\textwidth]{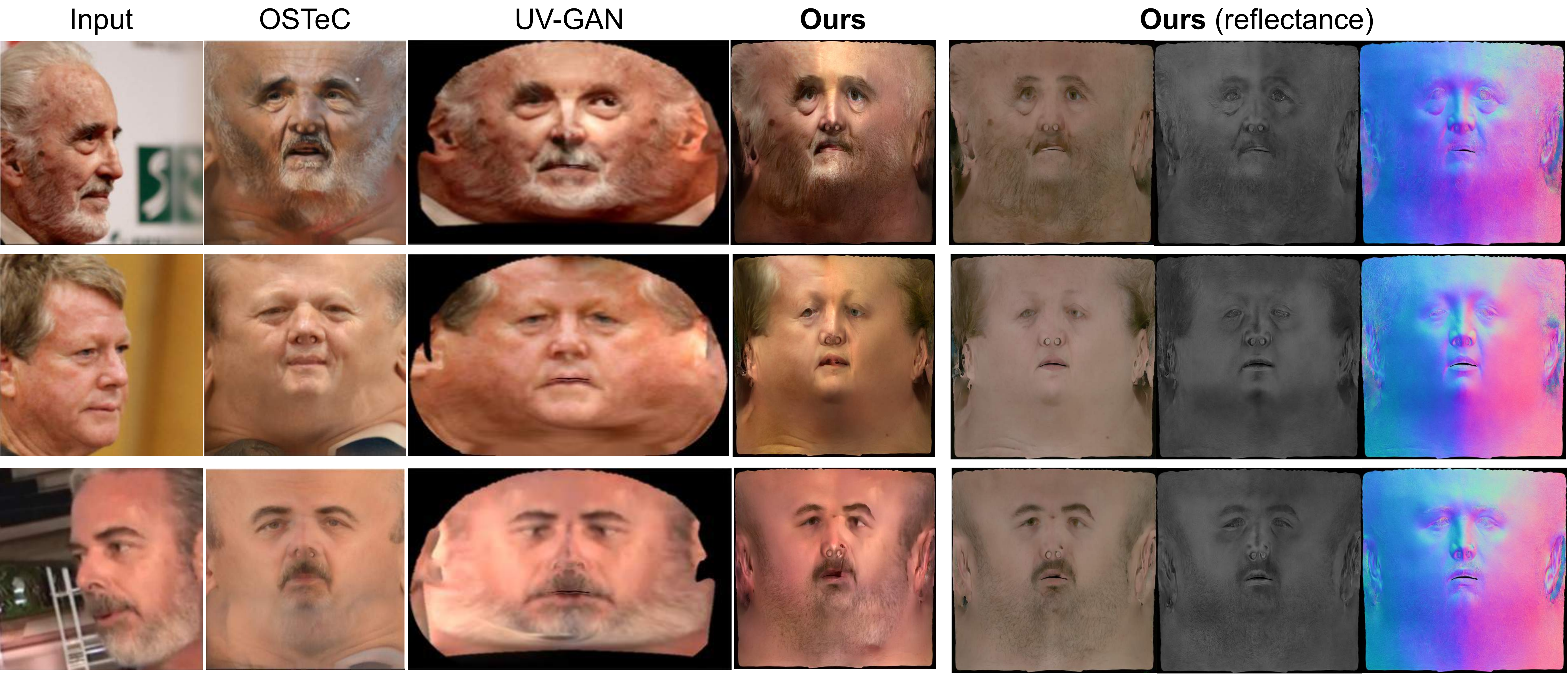}\\
\caption{Examples of texture completion by OSTeC~\cite{Gecer_ostec}, UV-GAN~\cite{Deng_UVGAN} and our method from ``in-the-wild'' images (each method uses a different UV topology). In contrast to ~\cite{Gecer_ostec,Deng_UVGAN} we additionally recover reflectance components for realistic rendering.}
\label{fig:uv_completion}
\end{figure}

Following~\cite{Gecer_ostec,Deng_UVGAN}, we evaluate our method on the task of texture completion using the Multi-PIE~\cite{multipie} subset of the UVDB dataset~\cite{Deng_UVGAN}. This consists of complete UV textures for 337 different identities, and corresponding 2D images of the faces from various camera poses. In accordance with~\cite{Gecer_ostec, Deng_UVGAN}, we use the last 137 subjects for evaluation (as the first 200 were used as training data in prior works). We perform texture completion with our diffusion-based approach for each different viewing angle and compare it with existing texture completion methods, namely CE~\cite{pathakCVPR16context}, UV-GAN~\cite{Deng_UVGAN} and OSTeC~\cite{Gecer_ostec}. We use the widely adopted Peak Signal-to-Noise Ratio (PSNR) and Structural Similarity Index (SSIM) metrics to compare the completed textures with the ground truth and report the results in Tab.~\ref{table:UVCompletion7views}. As can be seen, \textit{Relightify} outperforms the related methods in almost all settings, especially for challenging angles. A visual comparison with~\cite{Gecer_ostec, Deng_UVGAN} is provided in Fig.~\ref{fig:uv_completion}. Note that in contrast to CE~\cite{pathakCVPR16context} and UV-GAN~\cite{Deng_UVGAN}, our model was not trained on the Multi-PIE dataset.

\begin{table}
\footnotesize
\begin{center}
\resizebox{\linewidth}{!}{
\begin{tabular}{c|c|c|c|c|c}
\hline
Methods & Metric & $0^{\circ}$ & $\pm30^{\circ}$  & $\pm60^{\circ}$ & $\pm90^{\circ}$ \\
\hline
\multirow{2}{*} {CE~\cite{pathakCVPR16context}}  & PSNR & 23.03  & 21.93  & 20.27  & 19.63 \\
& SSIM & 0.920 & 0.892 & 0.888 & 0.718 \\ 
\hline
\multirow{2}{*} {UV-GAN~\cite{Deng_UVGAN}} & PSNR & 23.36  & 22.25  & 20.53  & 19.83 \\
& SSIM & 0.924 & 0.897 & 0.892 & 0.725 \\
\hline
\multirow{2}{*} {OSTeC~\cite{Gecer_ostec}}   & PSNR & 23.95 & 22.54  & 21.04  & 20.44  \\
 & SSIM & \textbf{0.928} & 0.902 & 0.898 & 0.746 \\
\hline
\multirow{2}{*} {Ours}   & PSNR & \textbf{26.00} & \textbf{24.73}  & \textbf{24.65}  &  \textbf{20.58} \\
 & SSIM & \textbf{0.928} & \textbf{0.916} & \textbf{0.917} & \textbf{0.874}\\
\hline
\end{tabular}
}
\end{center}
\caption{Quantitative comparison between \textit{Relightify} and ~\cite{pathakCVPR16context, Deng_UVGAN, Gecer_ostec} regarding UV texture completion on the MultiPIE dataset~\cite{multipie} for different viewing angles.}
\label{table:UVCompletion7views}
\end{table}

\subsection{Identity Preservation}
\begin{figure}[h]
  \centering
  \includegraphics[width=\linewidth, trim={2.5cm 0 1.9cm 0}, clip]{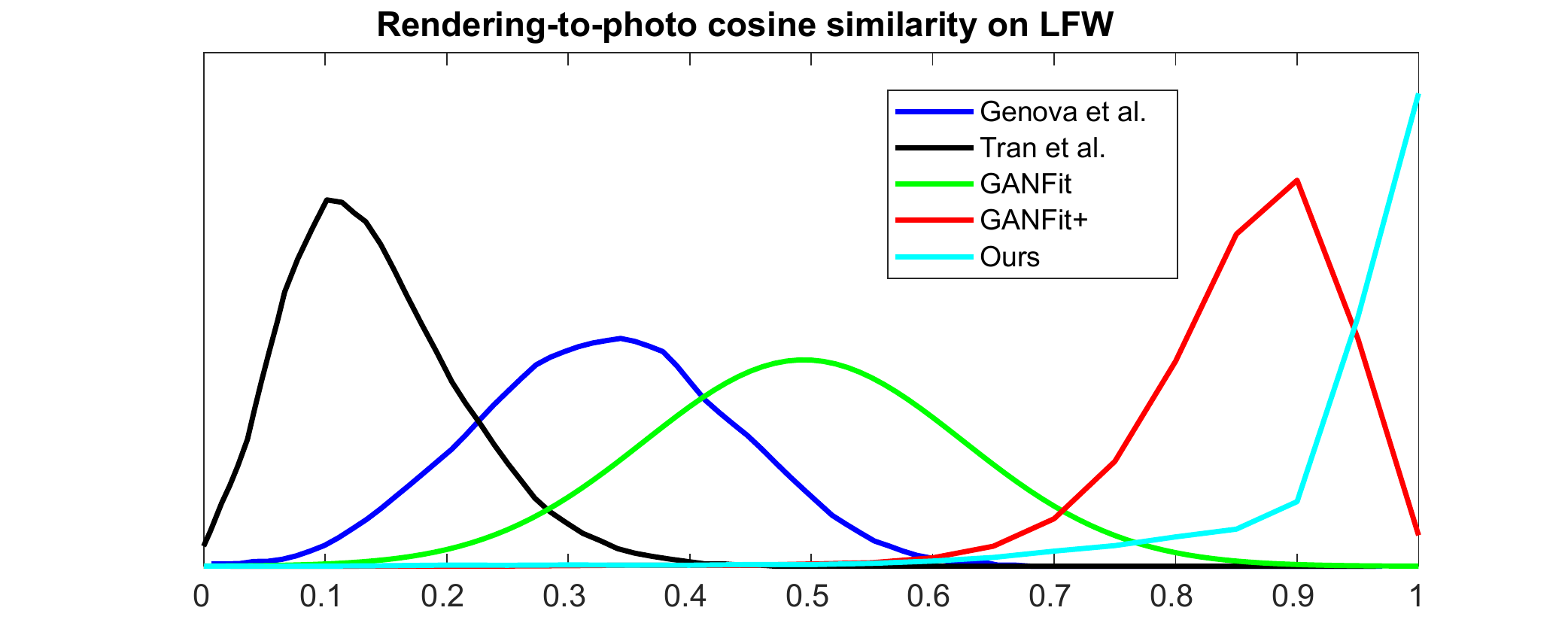}
  \caption{
    Quantitative evaluation of similarity scores on LFW~\cite{LFWTech},
    compared with prior work (\cite{genova2018unsupervised, tuan2017regressing, ganfit, fast-ganfit}), using VGG-Face~\cite{parkhi2015deep}.
    We show the cosine similarity distribution between ground truth and reconstruction.
  }
  \label{fig:exp_lfw_similarity}
\end{figure}

We perform quantitative evaluations of our method's ability to preserve the subject's identity,
by comparing the distribution of identity scores between the input image and rendered reconstruction, on the LFW dataset~\cite{LFWTech}, 
against prior work~\cite{ganfit, fast-ganfit, genova2018unsupervised, tuan2017regressing}.
Following the existing benchmark~\cite{fast-ganfit}, we evaluate our results using VGG-Face~\cite{parkhi2015deep}.
We present our analysis in Fig.~\ref{fig:exp_lfw_similarity},
measuring the distance between the input image and reconstruction for all subjects. Our method shows a significant improvement in similarity,
while also producing not just a facial texture, but a set of relightable reflectance textures.

\subsection{Reflectance Prediction}
To further assess our method on the task of facial reflectance prediction from monocular images, we use six test subjects with captured ground truth reflectance using a Light Stage \cite{ghosh2011multiview}, and compare \textit{Relightify} with the state-of-the-art method of AvatarMe++~\cite{lattas2021avatarme++}. More specifically, we apply both methods on 2D photos of these subjects and measure the PSNR of the recovered reflectance maps with respect to the ground truth maps. As shown in Tab.~\ref{tab:comp_ls}, our method produces significantly more accurate diffuse and specular albedos, while the normals closely match those of~\cite {lattas2021avatarme++}. This demonstrates our method's ability to better capture subject-specific details by directly leveraging texture information from the input image. Note that AvatarMe++ reconstructions are additionally conditioned on the 3DMM shape normals, which may explain a slight increase in the corresponding PSNR.

\begin{table}[h!]
\setlength{\tabcolsep}{2.5pt}
\footnotesize
\begin{center}
\begin{tabular}{c|c|c|c}
\multirow{2}{*}{} & PSNR & PSNR & PSNR\\
& (diffuse albedo) & (specular albedo) & (normals)\\
\hline
AvatarMe++~\cite{lattas2021avatarme++} & 18.30 & 19.77 & \textbf{27.26}\\
\hline
Ours & \textbf{22.47} & \textbf{27.17} & 26.69\\
\hline
\end{tabular}
\end{center}
\caption{Quantitative comparison of our method with \cite{lattas2021avatarme++} (results provided by authors). We calculate the average PSNR between the reconstructed and the ground truth reflectance maps for six subjects with ground truth, captured using a Light Stage \cite{ghosh2011multiview}.}
\label{tab:comp_ls}
\end{table}

\subsection{Experimentation with Inpainting Algorithms}
\begin{figure}[h]
\centering
\includegraphics[width=.47\textwidth]{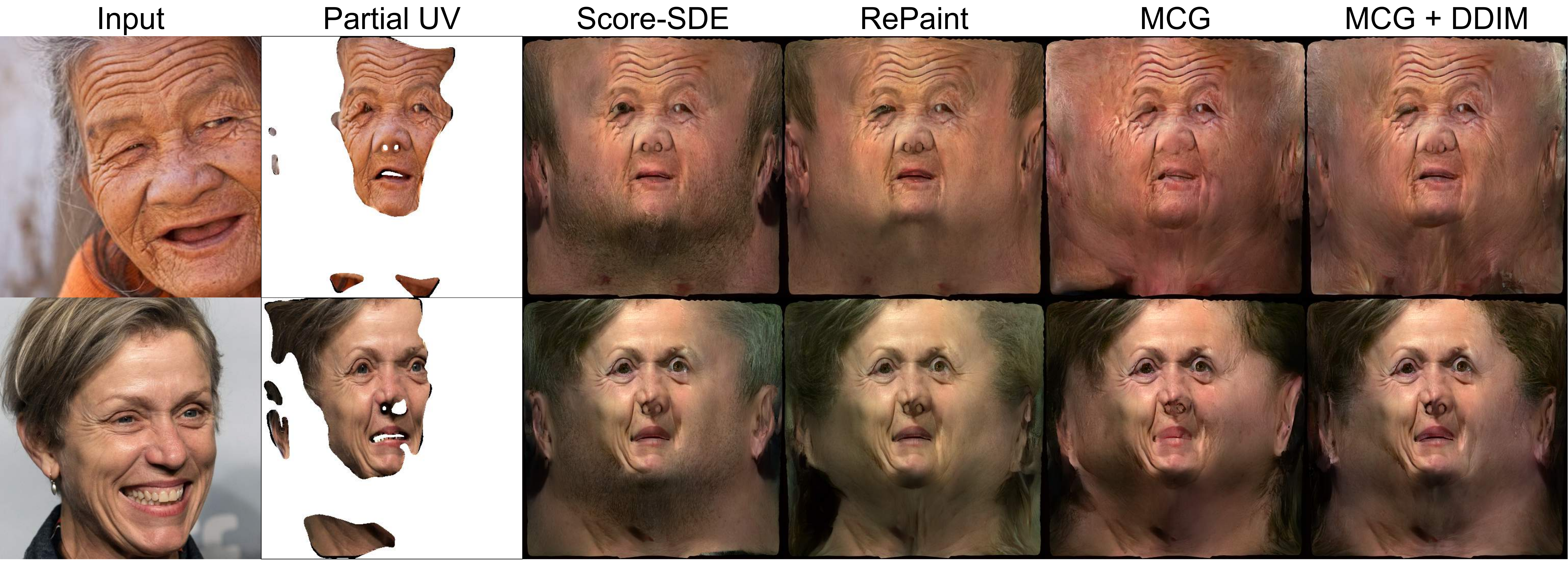}\\
\caption{Texture completion with our diffusion model using different inpainting algorithms~\cite{song2021Score-SDE, RePaint_2022_CVPR, chung2022MCG, ddim}. All algorithms are implemented on top of the same unconditionally trained diffusion model, and only the reverse sampling process is modified.}
\label{fig:inpaint_algo}
\end{figure}

Although we adopt the MCG~\cite{chung2022MCG} approach for our texture/reflectance diffusion model, we have experimented with different inpainting algorithms. We compare four of them in Fig.~\ref{fig:inpaint_algo} and Tab.~\ref{tab:algo_ls_comp}. We also provide the runtime for each algorithm in Tab.~\ref{tab:times}. The baseline method of Score-SDE~\cite{song2021Score-SDE}, which can be interpreted as Eq.~\ref{eq:inpainting_steps} without the gradient term, produces sub-optimal results, \ie the occluded areas are often inpainted in an inconsistent way with the observed ones, which is especially apparent in the texture (Fig.~\ref{fig:inpaint_algo}) and albedos (Tab.~\ref{tab:algo_ls_comp}). RePaint~\cite{RePaint_2022_CVPR} also produces unsatisfactory textures while at the same time increasing the reverse diffusion steps by a factor of $n$ (we use $n=10$ as suggested by the authors of ~\cite{RePaint_2022_CVPR}), which significantly affects the computational time. In contrast, MCG~\cite{chung2022MCG} preserves the original sampling length ($T=1000$ timesteps), hence being much more efficient. However, it is still slower than Score-SDE~\cite{song2021Score-SDE} since it requires the computation of a gradient for the manifold constraint at each step. In general, we found MCG~\cite{chung2022MCG} to perform better in most cases. To further strengthen the efficiency of our method, we have additionally incorporated the DDIM~\cite{ddim} acceleration technique in the MCG algorithm, which allows reducing the denoising steps to $N<T$ (we use $N=200$) without a significant drop in quality. In such case, our method can generate high-quality texture and reflectance assets from a partial UV texture in roughly 12 seconds, which is significantly faster than competing texture completion algorithms (\eg OSTeC~\cite{Gecer_ostec} requires around 10 minutes).

\begin{table}[h!]
\setlength{\tabcolsep}{2.5pt}
\footnotesize
\begin{center}
\begin{tabular}{c|c|c|c|c}
& Score-SDE & RePaint & MCG (Ours) & MCG (Ours) + DDIM\\
\hline
Time & 17 sec & 3 min & 1 min & 12 sec
\end{tabular}
\end{center}
\caption{Sampling time during texture completion and reflectance prediction for different inpainting algorithms~\cite{song2021Score-SDE, RePaint_2022_CVPR, chung2022MCG, ddim} (using an Nvidia RTX 2080 TI GPU).}
\label{tab:times}
\end{table}

\begin{table}[h!]
\setlength{\tabcolsep}{2.5pt}
\footnotesize
\begin{center}
\begin{tabular}{c|cc|cc|cc}
 {} & \multicolumn{2}{c|}{Diffuse Albedo} & \multicolumn{2}{c|}{Specular Albedo} & \multicolumn{2}{c}{Normals}\\
{} & PSNR & SSIM & PSNR & SSIM & PSNR & SSIM\\
\hline
Score-SDE & 20.80 & 0.808 & 26.69 & 0.845 & 26.86 & 0.784\\
\hline
RePaint & 20.08 & 0.813 & 26.65 & 0.848 & 27.27 & 0.801\\
\hline
MCG & 22.47 & 0.825 & 27.17 & 0.853 & 26.69 & 0.781\\
\hline
MCG + DDIM & 21.94 & 0.817 & 26.88 & 0.846 & 26.45 & 0.774\\
\hline
\end{tabular}
\end{center}
\caption{Comparison of inpainting algorithms~\cite{song2021Score-SDE, RePaint_2022_CVPR, chung2022MCG, ddim} applied on our diffusion model,
following the evaluation of Tab.~\ref{tab:comp_ls}.}
\label{tab:algo_ls_comp}
\end{table}

\section{Limitations}
Our method outperforms prior works on texture completion as well as the challenging task of reflectance prediction. This is accomplished by explicitly recovering information from the input image via inpainting. Nonetheless, similarly to related texture completion works~\cite{Gecer_ostec, Deng_UVGAN}, this also implies that the reconstructed texture is affected by the quality of the input image. Although the partial texture is first projected in our latent diffusion space by the perceptual encoder, a low resolution input may still degrade the quality of our result. In these cases, an upsampling network could be employed as in~\cite{lattas2020avatarme} to improve the resolution and details of the predicted UV maps. Also, despite its relatively large size, the employed dataset~\cite{papaioannou2022mimicme} may still under-represent some ethnic groups and lack diverse facial expressions, reducing accuracy in those cases. Incorporating diverse high-quality ground truth data with captured reflectance would significantly improve the performance. Finally, our method may also suffer by the ambiguity between albedo and illumination,
which is thoroughly described in TRUST \cite{feng2022towards}. In fact,
their proposed solution could be combined with our method in future work.

\section{Conclusion}
In this paper we introduced \textit{Relightify},
a method that achieves state-of-the-art facial texture completion
and facial reflectance acquisition,
from monocular ``in-the-wild'' images.
To achieve this, we train a latent diffusion model 
with multiple encoder-decoder networks,
on a synthetic facial texture and reflectance dataset,
and use a diffusion-based inpainting method on the masked UV textures.
Our results directly acquire the visible facial parts while also extrapolating to facial reflectance that exhibits a high likeness to the input image and can be trivially employed in commercial rendering applications.
\\
\textbf{Acknowledgements.} A. Lattas was partly funded by the EPSRC Fellowship DEFORM (EP/S010203/1). S. Zafeiriou and part of the research was funded by the EPSRC Fellowship DEFORM (EP/S010203/1) and EPSRC Project GNOMON (EP/X011364/1).

{\small
\bibliographystyle{ieee_fullname}
\bibliography{egbib}

\begin{thebibliography}{10}\itemsep=-1pt

\bibitem{DigitalEmily}
Oleg Alexander, Mike Rogers, William Lambeth, Jen-Yuan Chiang, Wan-Chun Ma,
  Chuan-Chang Wang, and Paul Debevec.
\newblock The digital emily project: Achieving a photorealistic digital actor.
\newblock {\em IEEE Computer Graphics and Applications}, 30(4):20--31, 2010.

\bibitem{hifi3dface2021tencentailab}
Linchao Bao, Xiangkai Lin, Yajing Chen, Haoxian Zhang, Sheng Wang, Xuefei Zhe,
  Di Kang, Haozhi Huang, Xinwei Jiang, Jue Wang, Dong Yu, and Zhengyou Zhang.
\newblock High-fidelity 3d digital human head creation from rgb-d selfies.
\newblock {\em ACM Transactions on Graphics}, 2021.

\bibitem{Blanz}
Volker Blanz and Thomas Vetter.
\newblock A morphable model for the synthesis of 3d faces.
\newblock In {\em Proceedings of the 26th Annual Conference on Computer
  Graphics and Interactive Techniques}, SIGGRAPH '99, page 187–194, USA,
  1999. ACM Press/Addison-Wesley Publishing Co.

\bibitem{LSFM}
James Booth, Anastasios Roussos, Stefanos Zafeiriou, Allan Ponniah, and David
  Dunaway.
\newblock A 3d morphable model learnt from 10,000 faces.
\newblock In {\em Proceedings of the IEEE Conference on Computer Vision and
  Pattern Recognition (CVPR)}, June 2016.

\bibitem{FaceWarehouse}
Chen Cao, Yanlin Weng, Shun Zhou, Yiying Tong, and Kun Zhou.
\newblock Facewarehouse: A 3d facial expression database for visual computing.
\newblock {\em IEEE Transactions on Visualization and Computer Graphics},
  20(3):413–425, 2014.

\bibitem{chen2019photo}
Anpei Chen, Zhang Chen, Guli Zhang, Kenny Mitchell, and Jingyi Yu.
\newblock Photo-realistic facial details synthesis from single image.
\newblock In {\em Proceedings of the IEEE International Conference on Computer
  Vision (ICCV)}, pages 9429--9439, 2019.

\bibitem{4DFAB}
Shiyang Cheng, Irene Kotsia, Maja Pantic, and Stefanos Zafeiriou.
\newblock 4dfab: A large scale 4d database for facial expression analysis and
  biometric applications.
\newblock In {\em Proceedings of the IEEE Conference on Computer Vision and
  Pattern Recognition (CVPR)}, June 2018.

\bibitem{ILVR_2021_ICCV}
Jooyoung Choi, Sungwon Kim, Yonghyun Jeong, Youngjune Gwon, and Sungroh Yoon.
\newblock Ilvr: Conditioning method for denoising diffusion probabilistic
  models.
\newblock In {\em Proceedings of the IEEE/CVF International Conference on
  Computer Vision (ICCV)}, pages 14367--14376, October 2021.

\bibitem{Come_Closer_2022_CVPR}
Hyungjin Chung, Byeongsu Sim, and Jong~Chul Ye.
\newblock Come-closer-diffuse-faster: Accelerating conditional diffusion models
  for inverse problems through stochastic contraction.
\newblock In {\em Proceedings of the IEEE/CVF Conference on Computer Vision and
  Pattern Recognition (CVPR)}, pages 12413--12422, June 2022.

\bibitem{chung2022MCG}
Hyungjin Chung, Byeongsu Sim, and Jong~Chul Ye.
\newblock Improving diffusion models for inverse problems using manifold
  constraints.
\newblock In {\em Advances in Neural Information Processing Systems}, 2022.

\bibitem{LYHM2017}
Hang Dai, Nick Pears, William A.~P. Smith, and Christian Duncan.
\newblock A 3d morphable model of craniofacial shape and texture variation.
\newblock In {\em Proceedings of the IEEE International Conference on Computer
  Vision (ICCV)}, 2017.

\bibitem{debevec2000acquiring}
Paul Debevec, Tim Hawkins, Chris Tchou, Haarm-Pieter Duiker, Westley Sarokin,
  and Mark Sagar.
\newblock Acquiring the reflectance field of a human face.
\newblock In {\em Proceedings of the 27th annual conference on Computer
  graphics and interactive techniques}, pages 145--156, 2000.

\bibitem{Deng_UVGAN}
Jiankang Deng, Shiyang Cheng, Niannan Xue, Yuxiang Zhou, and Stefanos
  Zafeiriou.
\newblock Uv-gan: Adversarial facial uv map completion for pose-invariant face
  recognition.
\newblock In {\em Proceedings of the IEEE Conference on Computer Vision and
  Pattern Recognition (CVPR)}, June 2018.

\bibitem{deng2019accurate}
Yu Deng, Jiaolong Yang, Sicheng Xu, Dong Chen, Yunde Jia, and Xin Tong.
\newblock Accurate 3d face reconstruction with weakly-supervised learning: From
  single image to image set.
\newblock In {\em IEEE Computer Vision and Pattern Recognition Workshops},
  2019.

\bibitem{diffusion_beats_gans}
Prafulla Dhariwal and Alexander Nichol.
\newblock Diffusion models beat gans on image synthesis.
\newblock In {\em Advances in Neural Information Processing Systems},
  volume~34, pages 8780--8794, 2021.

\bibitem{dib2021practical}
Abdallah Dib, Gaurav Bharaj, Junghyun Ahn, C{\'e}dric Th{\'e}bault, Philippe
  Gosselin, Marco Romeo, and Louis Chevallier.
\newblock Practical face reconstruction via differentiable ray tracing.
\newblock In {\em Computer Graphics Forum}, volume~40, pages 153--164. Wiley
  Online Library, 2021.

\bibitem{dib2021towards}
Abdallah Dib, Cedric Thebault, Junghyun Ahn, Philippe-Henri Gosselin, Christian
  Theobalt, and Louis Chevallier.
\newblock Towards high fidelity monocular face reconstruction with rich
  reflectance using self-supervised learning and ray tracing.
\newblock In {\em Proceedings of the IEEE/CVF International Conference on
  Computer Vision (ICCV)}, pages 12819--12829, 2021.

\bibitem{egger20203d}
Bernhard Egger, William~AP Smith, Ayush Tewari, Stefanie Wuhrer, Michael
  Zollhoefer, Thabo Beeler, Florian Bernard, Timo Bolkart, Adam Kortylewski,
  Sami Romdhani, et~al.
\newblock 3d morphable face models—past, present, and future.
\newblock {\em ACM Transactions on Graphics (TOG)}, 39(5):1--38, 2020.

\bibitem{Esser_2021_VQGAN}
Patrick Esser, Robin Rombach, and Bjorn Ommer.
\newblock Taming transformers for high-resolution image synthesis.
\newblock In {\em Proceedings of the IEEE/CVF Conference on Computer Vision and
  Pattern Recognition (CVPR)}, pages 12873--12883, June 2021.

\bibitem{feng2022towards}
Haiwen Feng, Timo Bolkart, Joachim Tesch, Michael~J Black, and Victoria
  Abrevaya.
\newblock Towards racially unbiased skin tone estimation via scene
  disambiguation.
\newblock In {\em Computer Vision--ECCV 2022: 17th European Conference, Tel
  Aviv, Israel, October 23--27, 2022, Proceedings, Part XIII}, pages 72--90.
  Springer, 2022.

\bibitem{Gecer_ostec}
Baris Gecer, Jiankang Deng, and Stefanos Zafeiriou.
\newblock Ostec: One-shot texture completion.
\newblock In {\em Proceedings of the IEEE/CVF Conference on Computer Vision and
  Pattern Recognition (CVPR)}, pages 7628--7638, June 2021.

\bibitem{gecer2020synthesizing}
Baris Gecer, Alexandros Lattas, Stylianos Ploumpis, Jiankang Deng, Athanasios
  Papaioannou, Stylianos Moschoglou, and Stefanos Zafeiriou.
\newblock Synthesizing coupled 3d face modalities by trunk-branch generative
  adversarial networks.
\newblock In {\em Computer Vision--ECCV 2020: 16th European Conference,
  Glasgow, UK, August 23--28, 2020, Proceedings, Part XXIX 16}, pages 415--433.
  Springer, 2020.

\bibitem{ganfit}
Baris Gecer, Stylianos Ploumpis, Irene Kotsia, and Stefanos Zafeiriou.
\newblock Ganfit: Generative adversarial network fitting for high fidelity 3d
  face reconstruction.
\newblock In {\em Proceedings of the IEEE Conference on Computer Vision and
  Pattern Recognition (CVPR)}, June 2019.

\bibitem{fast-ganfit}
Baris Gecer, Stylianos Ploumpis, Irene Kotsia, and Stefanos~P Zafeiriou.
\newblock Fast-ganfit: Generative adversarial network for high fidelity 3d face
  reconstruction.
\newblock {\em IEEE Transactions on Pattern Analysis and Machine Intelligence},
  2021.

\bibitem{genova2018unsupervised}
Kyle Genova, Forrester Cole, Aaron Maschinot, Aaron Sarna, Daniel Vlasic, and
  William~T Freeman.
\newblock Unsupervised training for 3d morphable model regression.
\newblock In {\em Proceedings of the IEEE Conference on Computer Vision and
  Pattern Recognition (CVPR)}, pages 8377--8386, 2018.

\bibitem{ghosh2011multiview}
Abhijeet Ghosh, Graham Fyffe, Borom Tunwattanapong, Jay Busch, Xueming Yu, and
  Paul Debevec.
\newblock Multiview face capture using polarized spherical gradient
  illumination.
\newblock {\em ACM Transactions on Graphics (TOG)}, 30(6):1--10, 2011.

\bibitem{goodfellow2020generative}
Ian Goodfellow, Jean Pouget-Abadie, Mehdi Mirza, Bing Xu, David Warde-Farley,
  Sherjil Ozair, Aaron Courville, and Yoshua Bengio.
\newblock Generative adversarial networks.
\newblock {\em Communications of the ACM}, 63(11):139--144, 2020.

\bibitem{gotardo2018practical}
Paulo Gotardo, J{\'e}r{\'e}my Riviere, Derek Bradley, Abhijeet Ghosh, and Thabo
  Beeler.
\newblock Practical dynamic facial appearance modeling and acquisition.
\newblock {\em ACM Transactions on Graphics (TOG)}, 37(6):1--13, 2018.

\bibitem{multipie}
Ralph Gross, Iain Matthews, Jeffrey Cohn, Takeo Kanade, and Simon Baker.
\newblock Multi-pie.
\newblock {\em Image and Vision Computing}, 28(5):807–813, May 2010.

\bibitem{ddpm_Ho}
Jonathan Ho, Ajay Jain, and Pieter Abbeel.
\newblock Denoising diffusion probabilistic models.
\newblock In {\em Advances in Neural Information Processing Systems},
  volume~33, pages 6840--6851, 2020.

\bibitem{LFWTech}
Gary~B. Huang, Manu Ramesh, Tamara Berg, and Erik Learned-Miller.
\newblock Labeled faces in the wild: A database for studying face recognition
  in unconstrained environments.
\newblock Technical Report 07-49, University of Massachusetts, Amherst, October
  2007.

\bibitem{Huynh_2018_Mesoscopic}
Loc Huynh, Weikai Chen, Shunsuke Saito, Jun Xing, Koki Nagano, Andrew Jones,
  Paul Debevec, and Hao Li.
\newblock Mesoscopic facial geometry inference using deep neural networks.
\newblock In {\em Proceedings of the IEEE Conference on Computer Vision and
  Pattern Recognition (CVPR)}, June 2018.

\bibitem{karrasprogressive}
Tero Karras, Timo Aila, Samuli Laine, and Jaakko Lehtinen.
\newblock Progressive growing of gans for improved quality, stability, and
  variation.
\newblock In {\em International Conference on Learning Representations}, 2018.

\bibitem{karras2019style}
Tero Karras, Samuli Laine, and Timo Aila.
\newblock A style-based generator architecture for generative adversarial
  networks.
\newblock In {\em Proceedings of the IEEE/CVF Conference on Computer Vision and
  Pattern Recognition (CVPR)}, pages 4401--4410, 2019.

\bibitem{karras2020analyzing}
Tero Karras, Samuli Laine, Miika Aittala, Janne Hellsten, Jaakko Lehtinen, and
  Timo Aila.
\newblock Analyzing and improving the image quality of stylegan.
\newblock In {\em Proceedings of the IEEE/CVF Conference on Computer Vision and
  Pattern Recognition (CVPR)}, pages 8110--8119, 2020.

\bibitem{kawar2022ddrm}
Bahjat Kawar, Michael Elad, Stefano Ermon, and Jiaming Song.
\newblock Denoising diffusion restoration models.
\newblock In {\em Advances in Neural Information Processing Systems}, 2022.

\bibitem{kong2021fast_sampling}
Zhifeng Kong and Wei Ping.
\newblock On fast sampling of diffusion probabilistic models.
\newblock In {\em ICML Workshop on Invertible Neural Networks, Normalizing
  Flows, and Explicit Likelihood Models}, 2021.

\bibitem{lattas2022practical}
Alexandros Lattas, Yiming Lin, Jayanth Kannan, Ekin Ozturk, Luca Filipi,
  Giuseppe~Claudio Guarnera, Gaurav Chawla, and Abhijeet Ghosh.
\newblock Practical and scalable desktop-based high-quality facial capture.
\newblock In {\em Computer Vision--ECCV 2022: 17th European Conference, Tel
  Aviv, Israel, October 23--27, 2022, Proceedings, Part VI}, pages 522--537.
  Springer, 2022.

\bibitem{lattas2020avatarme}
Alexandros Lattas, Stylianos Moschoglou, Baris Gecer, Stylianos Ploumpis,
  Vasileios Triantafyllou, Abhijeet Ghosh, and Stefanos Zafeiriou.
\newblock Avatarme: Realistically renderable 3d facial reconstruction
  "in-the-wild".
\newblock In {\em Proceedings of the IEEE/CVF Conference on Computer Vision and
  Pattern Recognition (CVPR)}, June 2020.

\bibitem{lattas2023fitme}
Alexandros Lattas, Stylianos Moschoglou, Stylianos Ploumpis, Baris Gecer,
  Jiankang Deng, and Stefanos Zafeiriou.
\newblock {FitMe}: Deep photorealistic {3D} morphable model avatars.
\newblock In {\em Proceedings of the IEEE/CVF Conference on Computer Vision and
  Pattern Recognition (CVPR)}, June 2023.

\bibitem{lattas2021avatarme++}
Alexandros Lattas, Stylianos Moschoglou, Stylianos Ploumpis, Baris Gecer,
  Abhijeet Ghosh, and Stefanos~P Zafeiriou.
\newblock Avatarme++: Facial shape and brdf inference with photorealistic
  rendering-aware gans.
\newblock {\em IEEE Transactions on Pattern Analysis and Machine Intelligence},
  2021.

\bibitem{Lee_2020_CVPR}
Gun-Hee Lee and Seong-Whan Lee.
\newblock Uncertainty-aware mesh decoder for high fidelity 3d face
  reconstruction.
\newblock In {\em IEEE/CVF Conference on Computer Vision and Pattern
  Recognition (CVPR)}, June 2020.

\bibitem{li2020learning}
Ruilong Li, Karl Bladin, Yajie Zhao, Chinmay Chinara, Owen Ingraham, Pengda
  Xiang, Xinglei Ren, Pratusha Prasad, Bipin Kishore, Jun Xing, et~al.
\newblock Learning formation of physically-based face attributes.
\newblock In {\em Proceedings of the IEEE/CVF Conference on Computer Vision and
  Pattern Recognition (CVPR)}, pages 3410--3419, 2020.

\bibitem{FLAME:SiggraphAsia2017}
Tianye Li, Timo Bolkart, Michael.~J. Black, Hao Li, and Javier Romero.
\newblock Learning a model of facial shape and expression from {4D} scans.
\newblock {\em ACM Transactions on Graphics, (Proc. SIGGRAPH Asia)},
  36(6):194:1--194:17, 2017.

\bibitem{li2017generative}
Yijun Li, Sifei Liu, Jimei Yang, and Ming-Hsuan Yang.
\newblock Generative face completion.
\newblock In {\em Proceedings of the IEEE Conference on Computer Vision and
  Pattern Recognition (CVPR)}, pages 3911--3919, 2017.

\bibitem{RePaint_2022_CVPR}
Andreas Lugmayr, Martin Danelljan, Andres Romero, Fisher Yu, Radu Timofte, and
  Luc Van~Gool.
\newblock Repaint: Inpainting using denoising diffusion probabilistic models.
\newblock In {\em Proceedings of the IEEE/CVF Conference on Computer Vision and
  Pattern Recognition (CVPR)}, pages 11461--11471, June 2022.

\bibitem{luo2021normalized}
Huiwen Luo, Koki Nagano, Han-Wei Kung, Qingguo Xu, Zejian Wang, Lingyu Wei,
  Liwen Hu, and Hao Li.
\newblock Normalized avatar synthesis using stylegan and perceptual refinement.
\newblock In {\em Proceedings of the IEEE/CVF Conference on Computer Vision and
  Pattern Recognition (CVPR)}, pages 11662--11672, 2021.

\bibitem{ma2007rapid}
Wan-Chun Ma, Tim Hawkins, Pieter Peers, Charles-Felix Chabert, Malte Weiss,
  Paul~E Debevec, et~al.
\newblock Rapid acquisition of specular and diffuse normal maps from polarized
  spherical gradient illumination.
\newblock {\em Rendering Techniques}, 2007(9):10, 2007.

\bibitem{marmoset}
{Marmoset LLC}.
\newblock Toolbag 4, version 4.051.

\bibitem{meng2022sdedit}
Chenlin Meng, Yutong He, Yang Song, Jiaming Song, Jiajun Wu, Jun-Yan Zhu, and
  Stefano Ermon.
\newblock {SDE}dit: Guided image synthesis and editing with stochastic
  differential equations.
\newblock In {\em International Conference on Learning Representations}, 2022.

\bibitem{papaioannou2022mimicme}
Athanasios Papaioannou, Baris Gecer, Shiyang Cheng, Grigorios Chrysos, Jiankang
  Deng, Eftychia Fotiadou, Christos Kampouris, Dimitrios Kollias, Stylianos
  Moschoglou, Kritaphat Songsri-In, et~al.
\newblock Mimicme: A large scale diverse 4d database for facial expression
  analysis.
\newblock In {\em Computer Vision--ECCV 2022: 17th European Conference, Tel
  Aviv, Israel, October 23--27, 2022, Proceedings, Part VIII}, pages 467--484.
  Springer, 2022.

\bibitem{parkhi2015deep}
Omkar~M. Parkhi, Andrea Vedaldi, and Andrew Zisserman.
\newblock Deep face recognition.
\newblock In {\em Proceedings of the British Machine Vision Conference (BMVC)},
  pages 41.1--41.12, September 2015.

\bibitem{pathakCVPR16context}
Deepak Pathak, Philipp Kr\"ahenb\"uhl, Jeff Donahue, Trevor Darrell, and Alexei
  Efros.
\newblock Context encoders: Feature learning by inpainting.
\newblock In {\em Proceedings of the IEEE Conference on Computer Vision and
  Pattern Recognition ({CVPR})}, 2016.

\bibitem{BaselFaceModel}
Pascal Paysan, Reinhard Knothe, Brian Amberg, Sami Romdhani, and Thomas Vetter.
\newblock A 3d face model for pose and illumination invariant face recognition.
\newblock In {\em 2009 Sixth IEEE International Conference on Advanced Video
  and Signal Based Surveillance}, pages 296--301, 2009.

\bibitem{riviere2020single}
J{\'e}r{\'e}my Riviere, Paulo Gotardo, Derek Bradley, Abhijeet Ghosh, and Thabo
  Beeler.
\newblock Single-shot high-quality facial geometry and skin appearance capture.
\newblock {\em ACM Transactions on Graphics (TOG)}, 39(4):81--1, 2020.

\bibitem{ldm_2022_CVPR}
Robin Rombach, Andreas Blattmann, Dominik Lorenz, Patrick Esser, and Bj\"orn
  Ommer.
\newblock High-resolution image synthesis with latent diffusion models.
\newblock In {\em Proceedings of the IEEE/CVF Conference on Computer Vision and
  Pattern Recognition (CVPR)}, pages 10684--10695, June 2022.

\bibitem{Saito_2017_CVPR}
Shunsuke Saito, Lingyu Wei, Liwen Hu, Koki Nagano, and Hao Li.
\newblock Photorealistic facial texture inference using deep neural networks.
\newblock In {\em Proceedings of the IEEE Conference on Computer Vision and
  Pattern Recognition (CVPR)}, July 2017.

\bibitem{robin2021noise_estimation}
Robin San{-}Roman, Eliya Nachmani, and Lior Wolf.
\newblock Noise estimation for generative diffusion models.
\newblock {\em CoRR}, abs/2104.02600, 2021.

\bibitem{sinha2021d2c}
Abhishek Sinha, Jiaming Song, Chenlin Meng, and Stefano Ermon.
\newblock D2c: Diffusion-denoising models for few-shot conditional generation.
\newblock {\em arXiv preprint arXiv:2106.06819}, 2021.

\bibitem{smith2020morphable}
William~AP Smith, Alassane Seck, Hannah Dee, Bernard Tiddeman, Joshua~B
  Tenenbaum, and Bernhard Egger.
\newblock A morphable face albedo model.
\newblock In {\em Proceedings of the IEEE/CVF Conference on Computer Vision and
  Pattern Recognition (CVPR)}, pages 5011--5020, 2020.

\bibitem{diffusion2015}
Jascha Sohl-Dickstein, Eric Weiss, Niru Maheswaranathan, and Surya Ganguli.
\newblock Deep unsupervised learning using nonequilibrium thermodynamics.
\newblock In {\em Proceedings of the 32nd International Conference on Machine
  Learning}, volume~37, pages 2256--2265, Lille, France, 07--09 Jul 2015.

\bibitem{ddim}
Jiaming Song, Chenlin Meng, and Stefano Ermon.
\newblock Denoising diffusion implicit models.
\newblock In {\em International Conference on Learning Representations}, 2021.

\bibitem{song2021Score-SDE}
Yang Song, Jascha Sohl-Dickstein, Diederik~P Kingma, Abhishek Kumar, Stefano
  Ermon, and Ben Poole.
\newblock Score-based generative modeling through stochastic differential
  equations.
\newblock In {\em International Conference on Learning Representations}, 2021.

\bibitem{Thies_2016_face2face}
Justus Thies, Michael Zollhofer, Marc Stamminger, Christian Theobalt, and
  Matthias Niessner.
\newblock Face2face: Real-time face capture and reenactment of rgb videos.
\newblock In {\em Proceedings of the IEEE Conference on Computer Vision and
  Pattern Recognition (CVPR)}, June 2016.

\bibitem{Tran_2019_CVPR}
Luan Tran, Feng Liu, and Xiaoming Liu.
\newblock Towards high-fidelity nonlinear 3d face morphable model.
\newblock In {\em Proceedings of the IEEE/CVF Conference on Computer Vision and
  Pattern Recognition (CVPR)}, June 2019.

\bibitem{tuan2017regressing}
Anh Tuan~Tran, Tal Hassner, Iacopo Masi, and G{\'e}rard Medioni.
\newblock Regressing robust and discriminative 3d morphable models with a very
  deep neural network.
\newblock In {\em Proceedings of the IEEE Conference on Computer Vision and
  Pattern Recognition (CVPR)}, pages 5163--5172, 2017.

\bibitem{vahdat2021score}
Arash Vahdat, Karsten Kreis, and Jan Kautz.
\newblock Score-based generative modeling in latent space.
\newblock In {\em Neural Information Processing Systems (NeurIPS)}, 2021.

\bibitem{Yamaguchi2018}
Shugo Yamaguchi, Shunsuke Saito, Koki Nagano, Yajie Zhao, Weikai Chen, Kyle
  Olszewski, Shigeo Morishima, and Hao Li.
\newblock High-fidelity facial reflectance and geometry inference from an
  unconstrained image.
\newblock {\em ACM Trans. Graph.}, 37(4), jul 2018.

\bibitem{3D_survey_sota}
Michael Zollhöfer, Justus Thies, Pablo Garrido, Derek Bradley, Thabo Beeler,
  Patrick Pérez, Marc Stamminger, Matthias Nießner, and Christian Theobalt.
\newblock State of the art on monocular 3d face reconstruction, tracking, and
  applications.
\newblock {\em Computer Graphics Forum}, 37:523--550, 2018.

\end{thebibliography}
}

\clearpage
\onecolumn
\appendix

\begin{center}
\textbf{\Large Relightify: Relightable 3D Faces from a Single Image via Diffusion Models}

\vspace{3pt}
\textbf{\Large (Supplementary Material)}
\vspace{30pt}
\end{center}

\section{Additional Qualitative Results}
\begin{figure*}[h]
\centering
\includegraphics[width=1.0\textwidth]{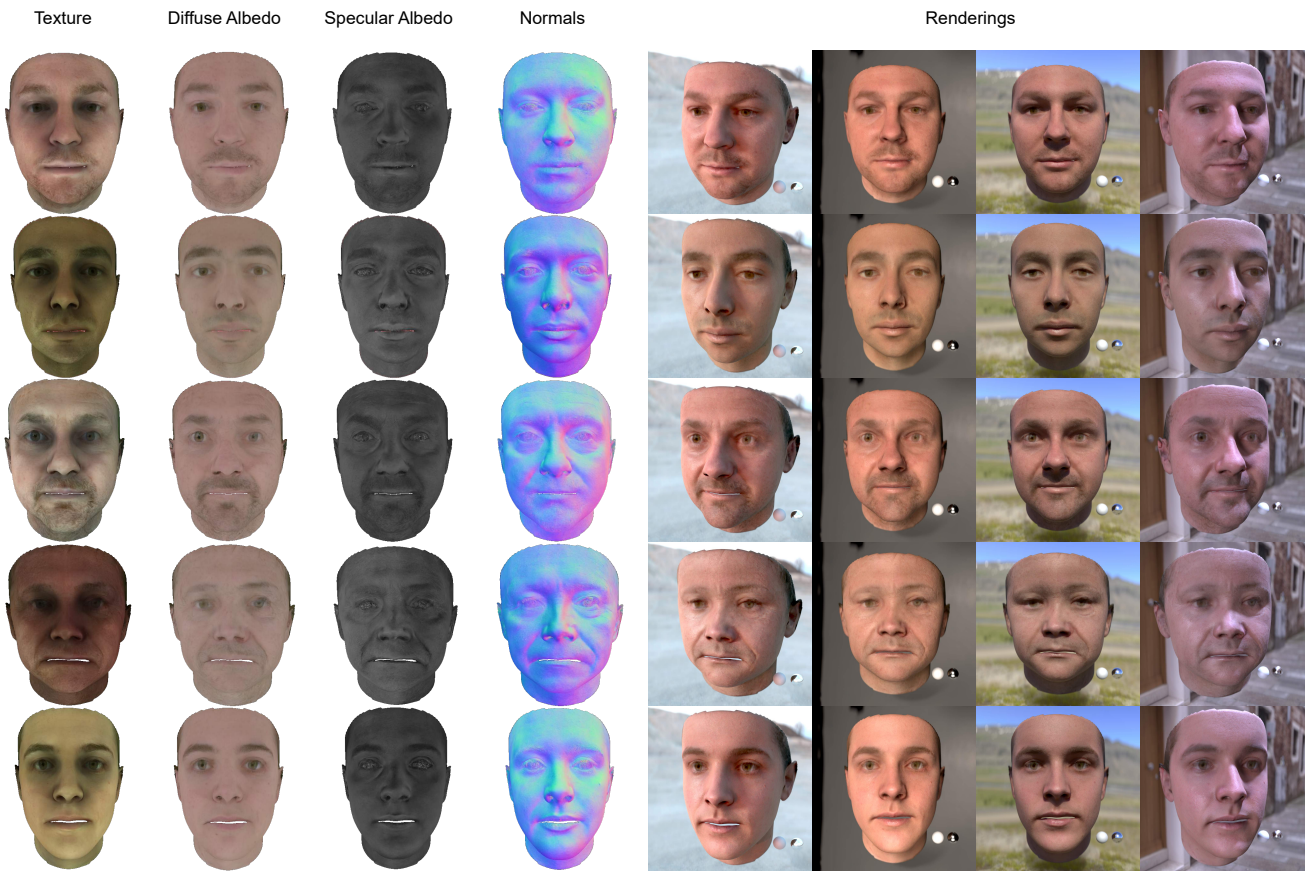}
\caption{Texture and reflectance maps sampled from our diffusion model. Left: We visualize the UV maps on top of arbitrary 3D shapes from a 3DMM~\cite{LSFM}. Right: Realistic renderings of the sampled reflectances on a commercial renderer.}
\label{fig:samples_supp}
\end{figure*}

\begin{figure*}[h]
\centering
\includegraphics[width=1.0\textwidth]{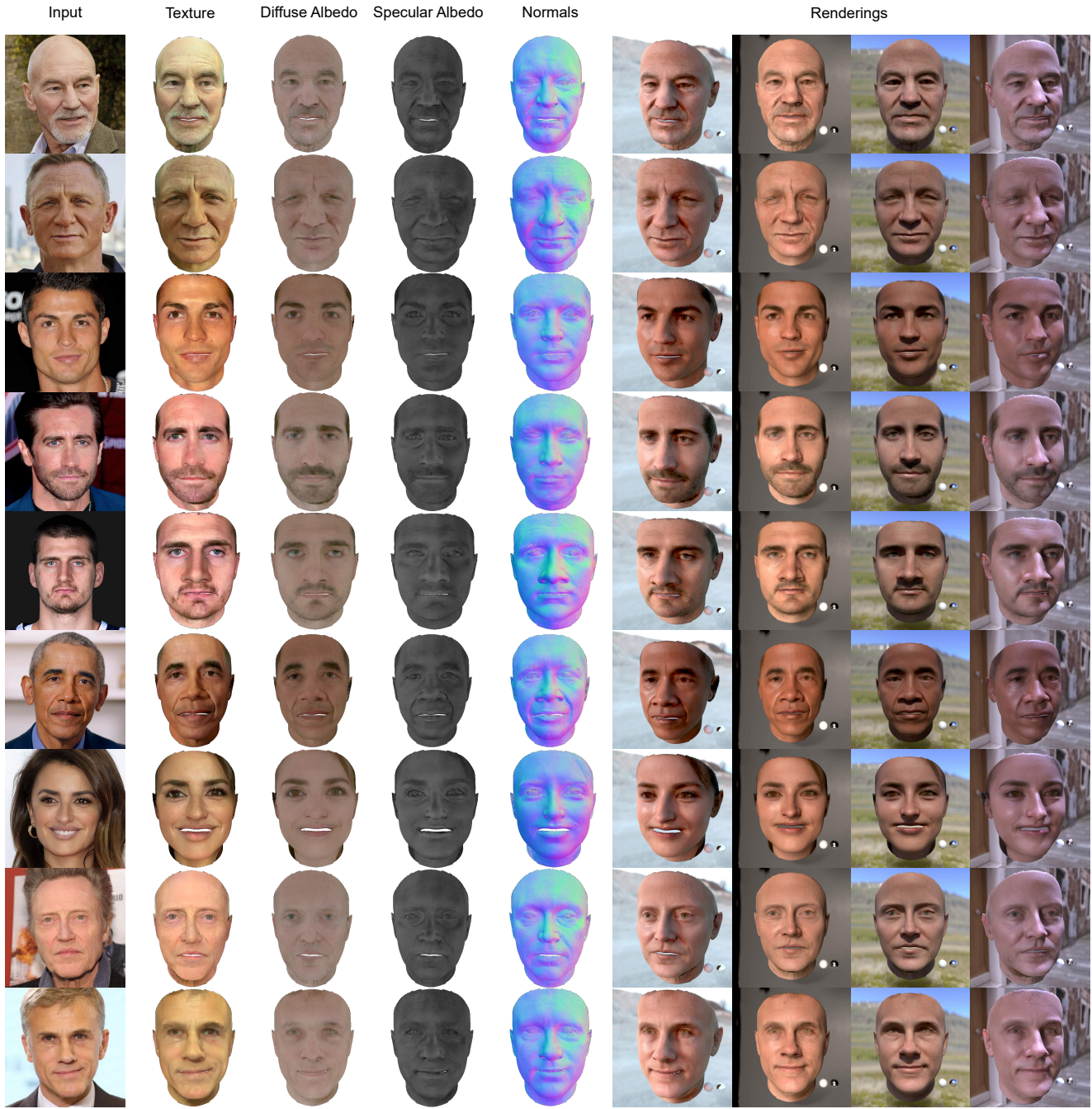}
\caption{3D reconstructions with recovered texture/reflectance maps by our method and renderings in different environments.}
\label{fig:recon_supp}
\end{figure*}

As discussed in the main paper, our method relies on a pre-trained diffusion model which serves as prior for guiding the inpainting procedure. To illustrate the generation capabilities of this model, we show an extension of Fig.~\ref{fig:samples} in Fig.~\ref{fig:samples_supp}, with a few more unconditionally generated (\ie based on the standard diffusion sampling process) samples. Besides the synthesized UV maps, we further apply the reflectance components on top of random 3D shapes drawn from the employed 3DMM and provide corresponding 3D renderings in different scenes. Finally, in Fig.~\ref{fig:recon_supp}, we show additional 3D reconstructions and renderings from monocular images with our proposed method.
\vspace{5mm}

\begin{figure*}[h!]
\centering
\includegraphics[width=\textwidth]{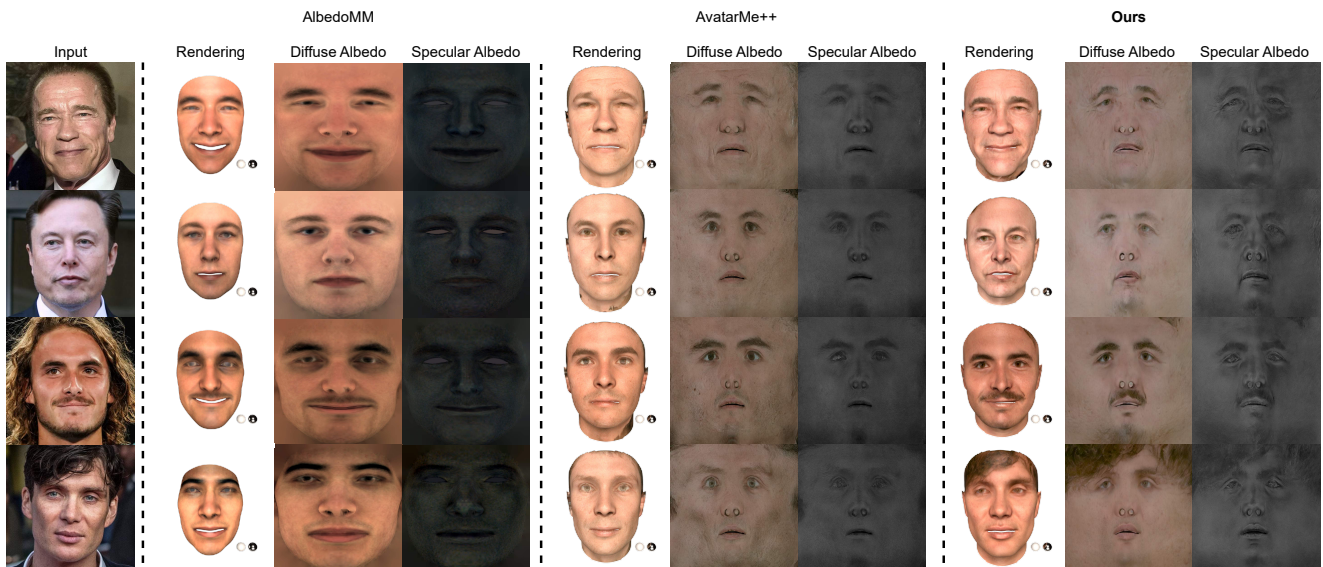}
\caption{Comparison with AvatarMe++ \cite{lattas2021avatarme++} and AlbedoMM \cite{smith2020morphable}.
For a fair comparison, we sample the per-vertex albedo values of AlbedoMM to a flat plane UV parameterization, and crop AvatarMe++ and Our results to the central facial area, since AlbedoMM is using a tigher facial crop.}
\label{fig:comparison_supp}
\end{figure*}

\section{Visual Comparison with Reflectance Reconstruction Methods}
In Fig.~\ref{fig:relightify_vs_amm_vs_av}, we have shown examples of rendered reconstructions by our method, as well as two related methods, namely AlbedoMM~\cite{smith2020morphable} and AvatarMe++~\cite{lattas2021avatarme++}, both of which recover the facial reflectance from singe images, similarly to our method. Here, we extend this comparison by adding the diffuse and specular albedos produced by these methods in Fig.~\ref{fig:comparison_supp}. Note that although AvatarMe++~\cite{lattas2021avatarme++} generates UV assets of higher resolution (4K by 6K), our method shows better detail preservation properties by directly capitalizing on the observed texture. Also, for the case of AlbedoMM~\cite{smith2020morphable},
which uses per-vertex albedo values,
we transform them into a flat plane UV parameterization for visualization purposes.

\section{Ablation Study on Texture Augmentation}
As mentioned in the paper, in order to enhance our model's generalization ability, we create arbitrary textures during training by re-rendering their delighted versions using random illumination environments. Preliminary experiments showed that this approach improves performance compared to using the original MimicMe textures. Here, we provide an evaluation of the model trained on the original textures, following the setting of Tab.~\ref{tab:comp_ls}.

\begin{table}[h!]
\setlength{\tabcolsep}{2.5pt}
\begin{center}
\begin{tabular}{c|c|c|c}
\multirow{2}{*}{} & PSNR & PSNR & PSNR\\
& (diffuse albedo) & (specular albedo) & (normals)\\
\hline
Ours (w/o aug.) & 22.07 & 26.98 & 26.63\\
\hline
Ours (w/ aug.) & \textbf{22.47} & \textbf{27.17} & \textbf{26.69}\\
\end{tabular}
\end{center}
\caption{Ablation study for UV texture augmentation (as in Tab.~\ref{tab:comp_ls}).}
\label{tab:ablation_aug}
\end{table}

\end{document}